\documentclass[sigconf]{acmart}
\usepackage{subfigure}
\usepackage{graphicx}
\usepackage{subcaption}
\usepackage{xcolor}
\usepackage{ulem}
\usepackage{multirow}
\usepackage{float}
\usepackage{placeins}
\usepackage{tabularx}
\usepackage{graphicx}
\usepackage{subcaption}
\usepackage{adjustbox} 
\usepackage{amsmath}
\usepackage{enumitem} 
\usepackage{algorithm}
\usepackage{algpseudocode}
\AtBeginDocument{%
  }

\setcopyright{acmlicensed}
\copyrightyear{2024}
\acmYear{2025}
\acmDOI{XXXXXXX.XXXXXXX}

\acmConference[Conference acronym 'XX]{Make sure to enter the correct
  conference title from your rights confirmation emai}{June 03--05,
  2018}{Woodstock, NY}
\acmISBN{978-1-4503-XXXX-X/18/06}

\begin{document}

\title{Top K Enhanced Reinforcement Learning Attacks on Heterogeneous Graph Node Classification}

\author{Honglin Gao}
\affiliation{%
  \department{School of Electrical and Electronic Engineering}
  \institution{Nanyang Technological University}
  \country{Singapore}}
\email{honglin001@e.ntu.edu.sg}

\author{Xiang Li}
\affiliation{%
  \department{School of Electrical and Electronic Engineering}
  \institution{Nanyang Technological University}
  \country{Singapore}}
\email{xiang002@e.ntu.edu.sg}

\author{Yajuan Sun}
\affiliation{%
  \institution{Singapore Institute of Manufacturing Technology (SIMTech)}
  \streetaddress{Agency for Science, Technology and Research (A*STAR)}
  \city{Singapore}
  \country{Singapore}
}
\email{yajuan.sun@simtech.a-star.edu.sg}

\author{Gaoxi Xiao}
\affiliation{%
  \department{School of Electrical and Electronic Engineering}
  \institution{Nanyang Technological University}
  \country{Singapore}}
\email{egxxiao@ntu.edu.sg}

\renewcommand{\shortauthors}{}

\begin{abstract}
Graph Neural Networks (GNNs) have achieved remarkable success on graph-structured data. However, their vulnerability to adversarial attacks in heterogeneous graph settings remains underexplored, where complex semantics and diverse node types pose unique challenges. In this paper, we propose HeteroKRLAttack(\uline{Hetero}geneous Graph Top-\uline{K} \uline{R}einforcement \uline{L}earning \uline{Attack}), a novel black-box evasion attack method tailored for heterogeneous graphs. It employs reinforcement learning to learn edge perturbation strategies through limited interaction with the victim model, without requiring access to model parameters and the complete adjacency matrix. To enhance attack efficiency, we introduce a Top-K action space reduction strategy that selects the most promising target nodes during inference. Our dual-policy architecture—TypeNet and ActionNet—captures relation-aware structural patterns and generates type-specific attack actions. Experiments on three real-world datasets (ACM, DBLP, IMDB) demonstrate that HeteroKRLAttack significantly degrades classification performance across various HGNNs (e.g., HAN, HGT, SimpleHGN), with up to 60\% accuracy drop under limited budget. Furthermore, the proposed method exhibits strong transferability across different model architectures and remains effective even against existing defense strategies designed for heterogeneous graphs. These results further reveal structural vulnerabilities in heterogeneous GNNs and offer valuable insights for the design of future adversarial defense mechanisms.
\end{abstract}






\maketitle

\section{Introduction}
Graph-structured data is ubiquitous in real-world applications, including social networks \citep{gnnandsocial, socialNet2, socialNet3}, biological networks \citep{gnnandbiological, biognn2, biognn3}, and knowledge graphs \citep{gnnandknowgraph, knowledgegraphgnn2,knowledgegraphgnn3}, where nodes and edges represent entities and their relationships. Unlike other types of structured data such as grids or sequences, graph data exhibits a non-Euclidean structure characterized by arbitrary connectivity and complex relational dependencies.  

Graphs can be broadly categorized into homogeneous and heterogeneous types. Homogeneous graphs consist of a single type of node and edge, making them relatively simple in structure. In contrast, heterogeneous graphs involve multiple types of nodes and edges, better capturing the rich semantics and structural complexity of real-world systems. For example, in an academic network, nodes may represent researchers, papers, and institutions, while edges denote authorship, citation, or affiliation relationships. Due to this expressive power, heterogeneous graphs are widely used in domains such as finance \citep{HGNNandFinance, hgnnFinance2, hgnnFinance3} and scientific literature analysis \citep{HGNNandCite, hgnnCitation2, hgnnCitation3}.

Node classification is a fundamental task in graph analysis, aiming to predict the category of nodes based on their attributes and the underlying graph structure. It plays a central role in many real-world applications. For example, in social networks, it helps infer user interests or community affiliations \citep{NodeclassificationinSocialNetworks, userinterest1}; in biological networks, it supports the prediction of gene functions or disease associations \citep{NodeClassificationAndBiologicalNet, predictionGene}. In the context of heterogeneous graphs, this task becomes more challenging due to the presence of multiple node and edge types, which introduce rich but complex semantic dependencies.

To effectively capture such heterogeneity, various Heterogeneous Graph Neural Networks (HGNNs) have been developed. Representative methods include HAN \citep{wang2019heterogeneous}, which employs hierarchical attention over metapaths to model semantic importance, and MAGNN \citep{MAGNN}, which enhances representation learning through intra- and inter-metapath aggregation. These methods perform well in heterogeneous graphs by generating useful node representations for classification.

Although Heterogeneous Graph Neural Networks (HGNNs) perform well in node classification, their robustness to adversarial attacks is still underexplored. In machine learning, small input perturbations can significantly degrade model performance—a vulnerability that is amplified in graph-based learning due to the propagation effect of message passing. This makes node classification models especially prone to attacks, raising concerns for safety-critical applications \citep{adversarialAttackDefinition}.

While adversarial attacks on homogeneous graphs have been widely studied \citep{AdversarialAttacksonGraphNeuralNetworks, adversarialattackGraph}, research on heterogeneous graphs remains limited \citep{HGAttack,adversarialattackHetero2}. The diversity of node and edge types adds structural and semantic complexity, making both attack and defense more difficult. In particular, black-box attacks—where the attacker lacks access to the model and full graph—are largely overlooked. These realistic scenarios, relying only on partial observations, demand more systematic investigation to develop effective black-box strategies for heterogeneous graph learning.

To address the above challenges, we propose a novel black-box evasion attack framework, HeteroKRLAttack(\uline{Hetero}geneous Graph Top-\uline{K} \uline{R}einforcement \uline{L}earning \uline{Attack}), tailored for heterogeneous graphs. Our method assumes no access to model architecture, parameters, or the complete graph structure, and relies solely on observable input-output behavior to degrade classification performance. We formulate the attack as a reinforcement learning problem, where the attacker acts as an agent and learns to select optimal edge or node modifications through policy gradient optimization. To handle the high complexity and large action space inherent in heterogeneous graphs, we introduce a Top-K node selection strategy based on KD-tree indexing, which significantly reduces the search space by focusing on the K most relevant nodes. This design improves both the efficiency and effectiveness of the attack, enabling the agent to identify impactful perturbations under constrained information.

In summary, our main contributions are as follows:
% \vspace{-5mm}
\begin{itemize}
    \item \textbf{Reinforcement learning black-box attack method}: We propose a reinforcement learning-based black-box attack method on heterogeneous graphs, attacking the model by only observing inputs and outputs.
    \item \textbf{Top-K method for action space reduction}: We utilize the Top-K method to effectively reduce the action space, improving search efficiency and attack effectiveness.
    \item \textbf{Experimental validation}: We conducted extensive experiments on multiple publicly available heterogeneous graph datasets; the results demonstrate that the proposed method significantly lowers the accuracy of node classification.
\end{itemize}

The paper is structured as follows: Section 2 reviews related work on heterogeneous neural networks and adversarial attacks on GNNs. Section 3 introduces preliminaries and formal definitions. In Section 4, we detail the proposed HeteroKRLAttack, including the reinforcement learning framework and the Top-K method for reducing the action space. Section 5 presents the experimental setup, results, and performance comparison with baseline methods, along with an ablation study. Section 6 concludes the paper and outlines future research directions. Our code has been open-sourced and is available at  https://anonymous.4open.science/r/HeteroKRL-Attack-4525.
% Matrices are denoted by uppercase letters, vectors by lowercase bold letters, scalars by lowercase letters, and sets by calligraphic letters; the graph is represented as GG.

\section{Related Work}
\subsection{Heterogeneous Graph Neural Network}
While homogeneous graphs have been extensively studied, heterogeneous graphs—with multiple node and edge types—pose greater challenges due to their complex semantic structures. Tasks like node classification become harder, as models must handle diverse relations and cross-type interactions.

To address these challenges, a range of Heterogeneous Graph Neural Networks (HGNNs) have been developed in recent years \citep{RHGNN,HGCN,MAGNN,RpHGNN}. Representative models include HAN \citep{wang2019heterogeneous}, which employs meta-path-based attention to selectively aggregate information along semantically meaningful paths; HGT \citep{heterogeneousgraphtransformer}, which leverages a transformer-based mechanism to dynamically handle type-specific message passing; and SimpleHGN \citep{SimpleHGN}, which simplifies the modeling of heterogeneity by embedding edge-type semantics directly into the message propagation process, resulting in improved computational efficiency without sacrificing performance. However, despite their effectiveness, the robustness of these models against adversarial threats has received relatively limited attention.

\vspace{-2mm}
\subsection{Adversarial Attacks on Graph Data}

In the field of adversarial attacks on graph data, recent research has highlighted the vulnerability of deep neural networks. While these networks excel in tasks like node classification \citep{RpHGNN} and link prediction \citep{MAGNN}, their robustness on graph data remains problematic \citep{Sun2022AdversarialAD}. Attack methods often involve subtle perturbations to graph structures or node features, misguiding models into incorrect classifications \citep{Goodfellow2014ExplainingAH,Szegedy2013IntriguingPO}. Among them, the most commonly used method in black-box attack strategy design is reinforcement learning. This method does not require knowledge of the target model's internal structure and relies solely on input-output behavior for the attack. For instance, a study by Dai et al. \citep{dai2018adversarial} demonstrated a reinforcement learning-based attack that only needs prediction labels from the target classifier. Another research by Ma et al. \citep{ma2021graph} in 2020 explored less detectable methods like graph rewiring, using deep reinforcement learning to devise effective attack strategies. These insights not only highlight the fragility of graph neural networks but also prompt further development of defensive measures for graph data.

Though research on the robustness of heterogeneous graphs is still in its early stages, there are several noteworthy contributions. Zhang et al.\citep{zhang2022robust} developed a robust HGNN framework called RoHe, which enhances the robustness of HGNNs against adversarial attacks by pruning malicious neighbors through an attention purifier. On the other hand, Zhao et al. \citep{zhao2024hgattack} introduced HGAttack, the first grey-box evasion attack method specifically designed for heterogeneous graphs, which generates perturbations by designing a new surrogate model and incorporating a semantic-aware mechanism.  

However, existing grey-box attack methods suffer from several limitations. They typically require access to gradient information and the complete graph structure, which is often unrealistic in practical scenarios. Their effectiveness also relies heavily on carefully designed surrogate models and semantic priors, limiting their adaptability to diverse or evolving heterogeneous graphs. Moreover, as gradient-based methods, they are susceptible to defense techniques such as gradient obfuscation and adversarial training. These drawbacks underscore the need for more flexible and robust black-box attack strategies.

\begin{figure}[t] % 使用figure环境
    \centering
    \includegraphics[width=\columnwidth]{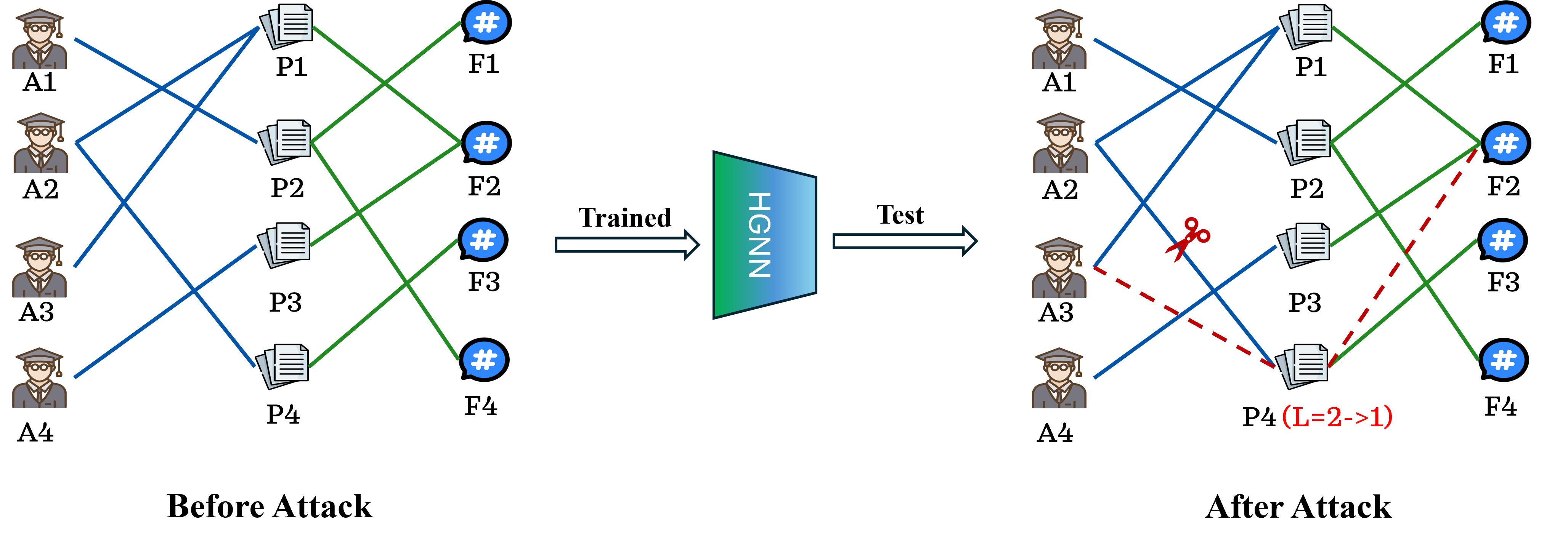} % 调整宽度以适应单栏
    \caption{Structural evasion attack on a heterogeneous GNN via edge perturbations}
    \label{fig:target attack example}
\end{figure}

\section{Preliminaries and Problem Definition}
\begin{figure*}[ht]
    \centering
    \includegraphics[width=0.7\linewidth]{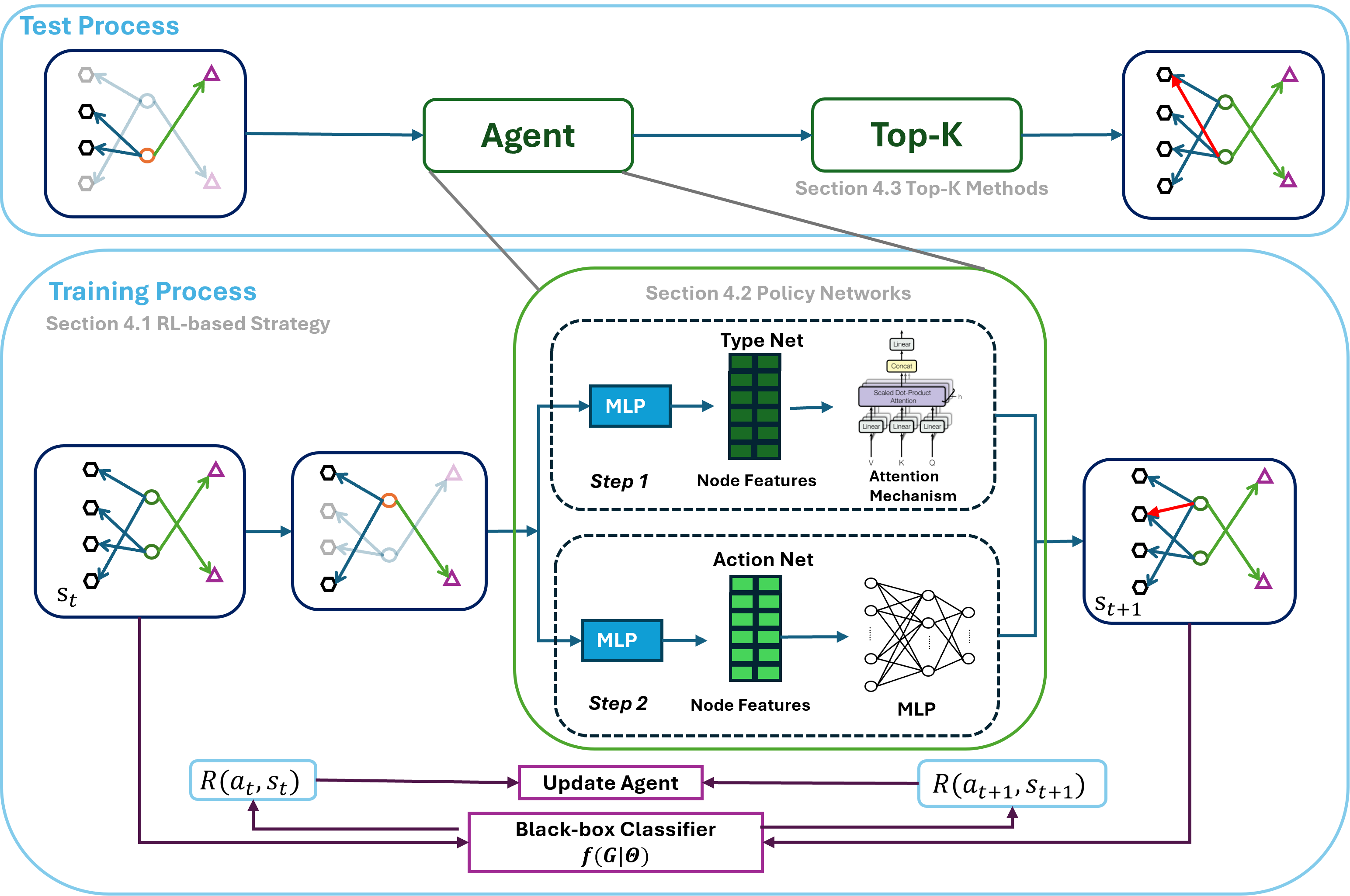} 
    \caption{Overall HeteroKRLAttack framework}
    \label{fig:Overall HeteroKRLAttack framework}
\end{figure*}
In this section, we present the preliminaries of backdoor attacks on heterogeneous graphs and formally define the problem setting. For clarity, Table~\ref{tab:symbols} provides a summary of the notations used throughout this section.

\begin{table}[t]
    \small
    \centering
    \begin{tabularx}{\linewidth}{@{}p{3.5cm} X@{}}
        \toprule
        \textbf{Symbol} & \textbf{Description} \\
        \midrule
        $G = (\mathcal{V}, \mathcal{E}, X)$ & Heterogeneous graph \\
        $\mathcal{V} = \{v_1, \ldots, v_n\}$ & Node set \\
        $\mathcal{E}$ & Edge set with types \\
        $X \in \mathbb{R}^{|\mathcal{V}| \times d}$ & Node features \\
        $\mathcal{T}$ & Node type set \\
        $\mathcal{R}$ & Edge type set \\
        $\mathcal{V}_t$ & Nodes of type $t$ \\
        $A_{t_a, t_b}$ & Adjacency matrix for type pair \\
        $(v_i, v_j, r_{t_a, t_b})$ & Typed edge \\
        $t_p$ & Primary type \\
        $\mathcal{T}_{\text{aux}}$ & Auxiliary types \\
        $v_{t_p}, v_{t_a}$ & Primary and auxiliary nodes \\
        $f$ & Trained model \\
        $y_{t_p}$ & True label of $v_{t_p}$ \\
        $\mathcal{L}$ & Loss function \\
        $\Delta \mathcal{E}$ & Edge modifications \\
        $G \oplus \Delta \mathcal{E}$ & Modified graph \\
        $c$ & Perturbation budget \\
        \bottomrule
    \end{tabularx}
    \caption{List of Symbols}
    \label{tab:symbols}
\end{table}

\subsection{Preliminaries}
\textbf{Definition 3.1 (Heterogeneous Graph)}  
A heterogeneous graph is defined as
$G = \{\mathcal{V}, \mathcal{E}, X\}$,
where $\mathcal{V} = \{v_1, v_2, \ldots, v_n\}$ is the node set, and $X \in \mathbb{R}^{|\mathcal{V}| \times d}$ is the node feature matrix with $d$ being the feature dimension.
The set of node types is defined as $\mathcal{T} = \{t_1, t_2, \ldots, t_T\}$, where each node $v \in \mathcal{V}$ belongs to exactly one type $t \in \mathcal{T}$. The subset of nodes of type $t$ is denoted as $\mathcal{V}_t \subseteq \mathcal{V}$, and its size is denoted by $\|\mathcal{V}_t\|$.
The set of edge types is denoted as
$\mathcal{R} = \left\{r_{t_a, t_b} \mid t_a, t_b \in \mathcal{T},\ t_a \neq t_b \right\}$
where each $r_{t_a, t_b} \in \mathcal{R}$ represents the relation from nodes of type $t_a$ to nodes of type $t_b$.

For each type pair $(t_a, t_b)$, we define a binary adjacency matrix $A_{t_a, t_b} \in \{0, 1\}^{\|\mathcal{V}_{t_a}\| \times \|\mathcal{V}_{t_b}\|}$, where $A_{t_a, t_b}(v_i, v_j) = 1$ indicates an edge from node $v_i \in \mathcal{V}_{t_a}$ to node $v_j \in \mathcal{V}_{t_b}$ under relation $r_{t_a, t_b}$.
Each edge in the graph is represented as a typed edge triplet:
$
(v_i, v_j, r_{t_a, t_b}), \quad \text{where } v_i \in \mathcal{V}_{t_a},\ v_j \in \mathcal{V}_{t_b},\ r_{t_a, t_b} \in \mathcal{R}
$
The overall edge set is defined as:
\[
\mathcal{E} = \bigcup_{r_{t_a, t_b} \in \mathcal{R}} \left\{ (v_i, v_j, r_{t_a, t_b}) \mid A_{t_a, t_b}(v_i, v_j) = 1 \right\}
\]
A graph is considered heterogeneous if it satisfies the condition:
$|\mathcal{T}| + |\mathcal{R}| > 2$

\textbf{Definition 3.2 (Meta-path)}
A metapath $m=t_1 \xrightarrow{r_1} t_2 \xrightarrow{r_2} \cdots \xrightarrow{r_{l}} t_{l+ 1}$ describes the pattern of connections between different node types $t_i \in \mathcal{T}$ via different edge types of $r_i \in \mathcal{R}$. The length of the metapath $M$ is the number of edges in the path.
For instance, in a bibliographic network, a metapath can be represented as ${\textit{Author}_1} \xrightarrow {\text{writes}} {\textit{Paper}_2} \xrightarrow  {\text {cites}} {\textit{Paper}_3} \xrightarrow{\text { written by }} {\textit{Author}_4}$, indicating a connection between two authors through papers they have written and cited.

\textbf{Definition 3.3 (Targeted evasion attack)}
A targeted attack focuses on a selected set of nodes, aiming to cause their predictions to become incorrect, regardless of the specific misclassified class. An evasion attack is performed after model training by modifying test-time inputs without changing the model parameters. A targeted evasion attack in the graph setting modifies only the graph structure—such as adding or removing edges—after training, in order to mislead the model on specific test nodes.

\textbf{Example 1}  As shown in Figure \ref{fig:target attack example}, we consider a heterogeneous graph with \textit{Author}, \textit{Paper}, and \textit{Field} nodes. A targeted evasion attack modifies the structure around a specific test paper node after training—e.g., by rewiring its links to authors or fields—without changing node features or model parameters. This perturbation leads the model to misclassify the selected node.

\textbf{Definition 3.4 (Primary type and auxiliary type)}  
In a heterogeneous graph, we define two key node types.
The \textit{primary type}, denoted as $t_p \in \mathcal{T}$, refers to the type of nodes that are the direct targets of classification or attack.
The \textit{auxiliary types}, denoted as $\mathcal{T}_{\mathrm{aux}}$, include all node types that are directly connected to nodes of the primary type via any edge type. Formally,
\begin{equation}
\begin{split}
\mathcal{T}_{\mathrm{aux}} = \Big\{ 
& t_a \in \mathcal{T} \setminus \{t_p\} \,\Big|\, 
\exists\, v_{t_p} \in \mathcal{V}_{t_p},\ 
v_{t_a} \in \mathcal{V}_{t_a},\ r \in \mathcal{R} : \\
& (v_{t_p}, v_{t_a}, r) \in \mathcal{E} \ 
\text{or} \ (v_{t_a}, v_{t_p}, r) \in \mathcal{E} 
\Big\}
\end{split}
\end{equation}
That is, $\mathcal{T}_{\mathrm{aux}}$ consists of all node types (excluding $t_p$) that have at least one direct connection to nodes of the primary type.

\textbf{Example 2.} In the bibliographic network shown in Figure~\ref{fig:target attack example}, where nodes are of types \textit{Author}, \textit{Paper}, and \textit{Field}, the \textit{Paper} type is considered the primary type if it is the target of attack, and the other types (\textit{Author} and \textit{Field}) are regarded as auxiliary types due to their direct interactions with \textit{Paper} nodes.

\subsection{Problem Definition}
Given a heterogeneous graph $G = (\mathcal{V}, \mathcal{E}, X)$, a trained node classification model $f$, and a set of target nodes $\mathcal{V}_{t_p}$ of primary type $t_p \in \mathcal{T}$, the goal of a targeted evasion attack is to perturb the graph structure after training such that the prediction of a specific test node $v_{t_p} \in \mathcal{V}_{t_p}$ becomes incorrect, regardless of the misclassified class.

We formulate the attack objective as maximizing the loss $\mathcal{L}$~\citep{Goodfellow-et-al-2016} on $v_{t_p}$ by modifying the graph edges:

\begin{equation}
\begin{aligned}
\max_{\Delta \mathcal{E}} \quad & \mathcal{L}\left(f\left(v_{t_p} \mid G \oplus \Delta \mathcal{E}\right), y_{t_p}\right) \\
\text{s.t.} \quad & \Delta \mathcal{E} \subseteq \left\{(v_{t_p}, v_{t_a}) \mid v_{t_p} \in \mathcal{V}_{t_p},\ v_{t_a} \in \mathcal{V}_{t_a},\ t_a \in \mathcal{T}_{\mathrm{aux}} \right\} \\
& |\Delta \mathcal{E}| \leq c
\end{aligned}
\label{eq:attack-objective}
\end{equation}

Here, $y_{t_p}$ is the true label of $v_{t_p}$, and $\Delta \mathcal{E}$ denotes the set of edge modifications. The operator $G \oplus \Delta \mathcal{E}$ indicates the graph after applying these changes. The constraint ensures that perturbations are restricted to edges between the primary type and auxiliary types. The parameter $c$ controls the total number of modifications. 

\section{HeteroKRLAttack Approach}
In this section, we propose HeteroKRLAttack, a reinforcement learning-based framework for structural attacks on heterogeneous graphs (see Fig.\ref{fig:Overall HeteroKRLAttack framework}). The framework consists of three core components. The first component (Section4.1) is the RL-based attack strategy, where we formulate the graph modification task as a Markov Decision Process and train an agent using the REINFORCE algorithm. The agent performs graph perturbations by selecting and connecting auxiliary nodes in a two-stage manner. The second component (Section~4.2) is the policy network design, which includes TypeNet and ActionNet for hierarchical decision-making. TypeNet uses multi-head attention to select auxiliary types, while ActionNet identifies specific nodes through residual semantic matching.

The third component (Section4.3) is the Top-K refinement mechanism applied during inference. After the agent selects an auxiliary node, we use its feature vector as a query in a KD-Tree to retrieve semantically similar neighbors, forming a local candidate pool. The final node is then selected based on its ability to flip the model prediction or maximize prediction discrepancy. Figure~\ref{fig:Overall HeteroKRLAttack framework} illustrates the full framework, aligning with Section 4.1 (RL Strategy), Section 4.2 (Policy Network), and Section 4.3 (Top-K Refinement).

Please refer to Appendix~\ref{appendix:algorithm_workflow} and Appendix~\ref{appendix:time_complexity} for the algorithm workflow and time complexity analysis.

\subsection{Reinforcement Learning-based Attack Strategy}
We formulate the graph modification problem as a reinforcement learning task, where the RL attack agent iteratively modifies the graph structure to fool the classifier $f(\cdot)$. The key components of our RL framework are defined as follows:

\begin{description}[leftmargin=1.8em, labelsep=0.5em]
  \item[State ($s$):] The state $s_t$ at time $t$ is defined as the tuple $(\mathcal{V}_{\text{obs}}, v_{t_p})$, where $v_{t_p}$ is the victim node, and $\mathcal{V}_{\text{obs}} \subseteq \mathcal{V}$ denotes the subset of nodes whose features and type information are accessible to the agent. The agent has no access to the edge structure and cannot observe how these nodes are connected.

  \item[Action ($a_t$):] The action involves two levels of granularity: at the coarse level, selecting an auxiliary type $t_a \in \mathcal{T}_{\mathrm{aux}}$; and at the fine level, choosing a specific auxiliary node $v_{t_a} \in \mathcal{V}_{t_a}$. The attacker then either removes or adds an edge between the primary node $v_{t_p}$ and the selected auxiliary node $v_{t_a}$.

  \item[Reward ($r$):] The reward measures the effectiveness of the graph modification in misleading the classifier. A higher reward is assigned when the model’s prediction on the modified graph $\hat{G}$ differs from that on the original graph $G$. The reward function is defined as:
  \begin{equation}
  r =
  \begin{cases}
  0, \quad \text{if the action is selecting an auxiliary type} \\ 
  \mathcal{L}(f(v_{t_p} \mid \hat{G}), y_{t_p}), \quad \text{if } f(v_{t_p} \mid \hat{G}) = y_{t_p} \\
  10, \quad \text{if } f(v_{t_p} \mid \hat{G}) \neq y_{t_p}
  \end{cases}
  \label{equ:whole function}
  \end{equation}
  where $\mathcal{L}(\cdot)$ denotes the cross-entropy loss between the prediction and true label $y_{t_p}$.

  \item[Policy ($\pi$):] The policy $\pi(a_t|s_t)$ defines the agent’s action-selection strategy based on the current state. It aims to maximize expected cumulative reward.

  \item[Termination:] The process stops under two conditions: (1) the attack succeeds, or (2) the number of modifications reaches the budget $c$.
\end{description}

To effectively implement this attack strategy, we chose the classic reinforcement learning algorithm 
 REINFORCE \citep{sutton1999policy}. The algorithm is a classic policy gradient algorithm for optimizing policies in reinforcement learning. It parameterizes the policy and updates its parameters using the Monte Carlo method to find the optimal policy. The process involves sampling trajectories from the current policy $\pi_\theta$ (consisting of states, actions, and rewards), then computing the return $R_t$ for each time step $t$ as the cumulative discounted reward:
\begin{equation}
R_t = \sum_{k=t}^T \gamma^{k-t} r_{k+1}
\end{equation}
where $\gamma$ is the discount factor. The policy parameters are updated via gradient ascent:
\begin{equation}
\theta \leftarrow \theta + \alpha \nabla_\theta \log \pi_\theta\left(a_t \mid s_t\right) R_t
\end{equation}
with $\alpha$ as the learning rate. The algorithm iteratively samples trajectories, computes returns, and updates the policy until convergence.

In the context of a heterogeneous graph attack, we employ a step-by-step approach to modify the graph structure. First, we select an auxiliary type $t_a \in \mathcal{T}_{\mathrm{aux}}$ based on the victim node $v_{t_p}$. Specifically, we choose an appropriate auxiliary type by considering the characteristics of the victim node and the available node-level information, without relying on edge connectivity.

The probability of selecting an auxiliary type $t_a$ given the state $s_t$ is denoted as $p\left(t_a \mid s_t\right)$. Once the auxiliary type is determined, we select a specific node $v_{t_a} \in \mathcal{V}_{t_a}$ belonging to the selected type. The probability of selecting this node given the current state and the chosen type is denoted as $p\left(v_{t_a} \mid s_t, t_a\right)$.

Finally, we add or delete an edge between the selected auxiliary node $v_{t_a}$ and the victim node $v_{t_p}$, thereby modifying the graph structure. This structured, two-step approach allows the agent to operate in a large action space more efficiently, improving the effectiveness of the attack.

Thus, the action selection probability is decomposed as:
\begin{equation}
   p\left(a_t \mid s_t\right) = p\left(t_a \mid s_t\right) \cdot p\left(v_{t_a} \mid s_t, t_a\right)
   \label{equ: generalize rl equation}
\end{equation}

The step-by-step approach is used to reduce the complexity of the action space in larger networks, where the extensive action space increases computational overhead. Breaking the process into manageable steps simplifies the calculations and improves computational efficiency, reducing time and resource consumption. Figure \ref{fig:Overall HeteroKRLAttack framework} illustrates the overall method.

\subsection{Policy Network}
We design two policy networks to estimate the distributions in Equation~(\ref{equ: generalize rl equation}). The first, TypeNet, selects an auxiliary type using a multi-head attention mechanism. Specifically, the victim node embedding is used as the query vector, while each key and value vector corresponds to the average embedding of all nodes belonging to a particular type. The multi-head attention computes compatibility between the victim node and each node type. The resulting attention scores are passed through a softmax function to obtain the probability distribution over auxiliary types:
{\small
\begin{align}
p\left(t_a \mid s_t\right) =\ & \operatorname{softmax}\Bigg( \mathbf{W}_{\text{out}} \Big( \sum_i \alpha_i \mathbf{v}_i \Big) + \mathbf{b}_{\text{out}} \Bigg) \label{eq:typeselect_prob} \\
\text{where } \alpha_i =\ & \frac{\exp(\mathbf{q} \cdot \mathbf{k}_i / \sqrt{d_k})}{\sum_j \exp(\mathbf{q} \cdot \mathbf{k}_j / \sqrt{d_k})} \label{eq:attention_score}
\end{align}
}
Here, $\mathbf{q}$ is the query vector for the victim node, and $\mathbf{k}_i$, $\mathbf{v}_i$ are the average key and value embeddings of type $t_i \in \mathcal{T}_{\mathrm{aux}}$. This attention mechanism enables the model to focus on semantically relevant auxiliary types in a compact and interpretable way.

ActionNet is designed to select a specific auxiliary node $v_{t_a} \in \mathcal{V}_{t_a}$ from the chosen auxiliary type $t_a \in \mathcal{T}_{\mathrm{aux}}$ under the current state $s_t$. To achieve this, the embeddings of the victim node $v_{t_p}$ and candidate auxiliary nodes $v_{t_a}$ are first projected into a shared representation space through linear transformation, layer normalization, and ReLU activation.

For each candidate node, its representation is concatenated with two contextual vectors: the average embedding of all candidate nodes from type $t_a$, denoted as $\overline{\mathbf{h}}_{t_a}$, and the embedding of the victim node $v_{t_p}$, denoted as $\mathbf{h}_{v_{t_p}}$. The concatenated features are passed through a multi-layer residual MLP (ResNet) to model the interaction between node-level and type-level semantics. Finally, a matching layer computes the dot product between the ResMLP output and the victim node embedding, followed by a softmax function to produce the action probability distribution:
{\small
\begin{align}
p(v_{t_a} \mid s_t, t_a) =
\operatorname{softmax} \left( 
    \text{Match}\left( 
        \text{ResNet}\left(
            [\mathbf{h}_{v_{t_a}};\, \overline{\mathbf{h}}_{t_a};\, \mathbf{h}_{v_{t_p}}]
        \right), 
    \mathbf{h}_{v_{t_p}} \right)
\right)
\end{align}
}
Here, $\mathbf{h}_{v_{t_a}}$ denotes the embedding of a candidate auxiliary node, $\overline{\mathbf{h}}_{t_a}$ represents the mean embedding of all candidates of type $t_a$, and $\mathbf{h}_{v_{t_p}}$ is the embedding of the victim node.

\subsection{Top-K Method for Reducing Action Space}
We observed that for large-scale networks, even when actions are divided into several parts according to auxiliary types, the action space remains vast, leading to inefficient search processes. To address this challenge, inspired by Google DeepMind's success in handling large action spaces \citep{2015deepRL}, we introduced a Top-K refinement method during the testing phase. This strategy identifies a small set of semantically similar candidate nodes around the one selected by the RL agent, thereby significantly reducing the number of actions to consider and enhancing decision-making efficiency and accuracy. To implement this process efficiently, we adopt the KD-Tree (K-Dimensional Tree) \citep{kdtree} as the underlying structure.

KD-Tree is an efficient structure for organizing k-dimensional spatial data, commonly used in tasks such as nearest neighbor search and range queries. It is built by recursively selecting median points along alternating dimensions to partition the space. During query, the tree is traversed based on the target dimension, and backtracking is used to explore other branches as needed, enabling efficient search in high-dimensional spaces.

In our framework, once the RL agent has selected an auxiliary node $v_{t_a}$, we take its raw feature vector as the query point. A KD-Tree is constructed from all candidate auxiliary nodes of the same type using their original features. By querying with the selected node’s feature, the KD-Tree retrieves its $K$ nearest neighbors in feature space, forming a refined local candidate pool.

From this Top-K set, we select the final node for graph modification based on the following criteria:
\textbf{(i)} If any node leads to a prediction flip by the target model, one of them is selected at random;
\textbf{(ii)} Otherwise, we choose the node that maximizes the discrepancy between predictions on the modified and original graph, as defined in Equation~(\ref{equ:whole function}).

%%%%%%%%%%%%%%%%%%%%%%%%%%%%%%%%%%%%%%%%%%%%%%%%%%%%%%%%%
\begin{table*}[ht]
\centering
\scriptsize
\setlength{\tabcolsep}{12pt}
\begin{tabular}{c|c|l|c|c|c|c|c|c}
\hline
\multirow{2}{*}{\textbf{Victim Model}} 
  & \multirow{2}{*}{\textbf{Budget}}  
  & \multirow{2}{*}{\textbf{Attack Method}} 
  & \multicolumn{2}{c|}{\textbf{ACM}}
  & \multicolumn{2}{c|}{\textbf{DBLP}}
  & \multicolumn{2}{c|}{\textbf{IMDB}} \\ 
\cline{4-9}
  & & & Micro-F1 & Macro-F1 & Micro-F1 & Macro-F1 & Micro-F1 & Macro-F1 \\ 
\hline
\multirow{16}{*}{SimpleHGN}
  & 0 & None      & 0.9108 & 0.9143 & 0.9655 & 0.9647 & 0.6014 & 0.6004 \\ 
\cline{2-9}
  & \multirow{5}{*}{1}
    & Random     & 0.8836 & 0.8861 & 0.8793 & 0.8790 & 0.5897 & 0.5840 \\
  &               & FGA        & 0.9678 & 0.9658 & 0.8990 & 0.8966 & 0.5454 & 0.5437 \\
  &               & HANAttack  & 0.8837 & 0.8835 & 0.9557 & 0.9552 & 0.5927 & 0.5884 \\
  &               & HeteroRL(w/o Top-K)   & $0.3663^\dagger$ & $0.3292^\dagger$ & $0.4187^\dagger$ & $0.4019^\dagger$ & $0.5431^\dagger$ & $0.5425^\dagger$ \\
  &               & HeteroKRL  & \textbf{0.1708} & \textbf{0.1908} & \textbf{0.3350} & \textbf{0.3471} & \textbf{0.3916} & \textbf{0.3955} \\
\cline{2-9}
  & \multirow{5}{*}{3}
    & Random     & 0.8564 & 0.8583 & 0.7906 & 0.7911 & 0.5687 & 0.5597 \\
  &               & FGA        & 0.8812 & 0.8421 & 0.5763 & 0.5721 & 0.3590 & 0.3600 \\
  &               & HANAttack  & 0.8069 & 0.7936 & 0.7586 & 0.7496 & 0.5617 & 0.5436 \\
  &               & HeteroRL(w/o Top-K)   & $0.3490^\dagger$ & $0.2695^\dagger$ & $0.3817^\dagger$ & $0.3477^\dagger$ & $0.2308^\dagger$ & $0.1910^\dagger$ \\
  &               & HeteroKRL  & \textbf{0.1485} & \textbf{0.1280} & \textbf{0.1379} & \textbf{0.1356} & \textbf{0.0979} & \textbf{0.0962} \\
\cline{2-9}
  & \multirow{5}{*}{5}
    & Random     & 0.8267 & 0.8304 & 0.7143 & 0.7148 & 0.5618 & 0.5545 \\
  &               & FGA        & 0.7921 & 0.7060 & 0.4212 & 0.4174 & 0.2681 & 0.2707 \\
  &               & HANAttack  & 0.7351 & 0.6669 & 0.5369 & 0.5137 & 0.5571 & 0.5365 \\
  &               & HeteroRL(w/o Top-K)   & $0.2450^\dagger$ & $0.2109^\dagger$ & $0.3694^\dagger$ & $0.3344^\dagger$ & $0.2098^\dagger$ & $0.1972^\dagger$ \\
  &               & HeteroKRL  & \textbf{0.0668} & \textbf{0.0672} & \textbf{0.0812} & \textbf{0.0818} & \textbf{0.0746} & \textbf{0.0747} \\
\cline{1-9}
\multirow{16}{*}{HGT}
  & 0 & None      & 0.9158 & 0.9170 & 0.9557 & 0.9539 & 0.6503 & 0.6534 \\ 
\cline{2-9}
  & \multirow{5}{*}{1}
    & Random     & 0.8837 & 0.8823 & 0.8522 & 0.8492 & 0.6527 & 0.6527 \\
  &               & FGA        & 0.9406 & 0.9345 & 0.8325 & 0.8324 & $0.5548\dagger$ & $0.5581\dagger$ \\
  &               & HANAttack  & 0.8465 & 0.8404 & 0.8793 & 0.8834 & 0.5921 & 0.5906 \\
  &               & HeteroRL(w/o Top-K)   & $0.6832^\dagger$ & $0.5972^\dagger$ & $0.6749^\dagger$ & $0.6873^\dagger$ & 0.5804 & 0.5606 \\
  &               & HeteroKRL  & \textbf{0.5421} & \textbf{0.4721} & \textbf{0.5369} & \textbf{0.5503} & \textbf{0.2797} & \textbf{0.2895} \\
\cline{2-9}
  & \multirow{5}{*}{3}
    & Random     & 0.8146 & 0.8359 & 0.7463 & 0.7429 & 0.5897 & 0.5833 \\
  &               & FGA        & 0.9035 & 0.8866 & $0.5271\dagger$ & $0.5209\dagger$ & $0.3473\dagger$ & $0.3555\dagger$ \\
  &               & HANAttack  & 0.6831 & 0.6193 & 0.7266 & 0.7099 & 0.5734 & 0.5652 \\
  &               & HeteroRL(w/o Top-K)   & $0.4381^\dagger$ & $0.3856^\dagger$ & 0.6453 & 0.6405 & 0.4965 & 0.4351 \\
  &               & HeteroKRL  & \textbf{0.2030} & \textbf{0.1585} & \textbf{0.1822} & \textbf{0.1839} & \textbf{0.0466} & \textbf{0.0489} \\
\cline{2-9}
  & \multirow{5}{*}{5}
    & Random     & 0.8317 & 0.8212 & 0.6871 & 0.6780 & 0.5804 & 0.5815 \\
  &               & FGA        & 0.8094 & 0.7753 & $0.3768\dagger$ & $0.3678\dagger$ & $0.2681\dagger$ & $0.2713\dagger$ \\
  &               & HANAttack  & 0.7129 & 0.6401 & 0.5000 & 0.4740 & 0.5594 & 0.5531 \\
  &               & HeteroRL(w/o Top-K)   & $0.2946^\dagger$ & $0.2567^\dagger$ & 0.5221 & 0.5148 & 0.4219 & 0.4195 \\
  &               & HeteroKRL  & \textbf{0.0965} & \textbf{0.0789} & \textbf{0.0812} & \textbf{0.0829} & \textbf{0.0186} & \textbf{0.0196} \\
\cline{1-9}
\multirow{16}{*}{HAN}
  & 0 & None     & 0.9109 & 0.9112 & 0.9409 & 0.9383 & 0.6294 & 0.6280 \\ 
\cline{2-9}
  & \multirow{5}{*}{1}
    & Random     & 0.8639 & 0.8568 & 0.7882 & 0.7834 & 0.6154 & 0.6110 \\
  &               & FGA        & 0.7946 & 0.7203 & 0.6995 & 0.6744 & 0.5618 & 0.5624 \\
  &               & HANAttack  & 0.6361 & 0.5065 & 0.9039 & 0.9008 & 0.5128 & 0.5045 \\
  &               & HeteroRL(w/o Top-K)   & $0.5009^\dagger$ & $0.2616^\dagger$ & $0.3005^\dagger$ & $0.2172^\dagger$ & $0.4709^\dagger$ & $0.4630^\dagger$ \\
  &               & HeteroKRL  & \textbf{0.3144} & \textbf{0.1890} & \textbf{0.2512} & \textbf{0.1613} & \textbf{0.0932} & \textbf{0.0928} \\
\cline{2-9}
  & \multirow{5}{*}{3}
    & Random     & 0.8218 & 0.8088 & 0.6133 & 0.6041 & 0.5921 & 0.5858 \\
  &               & FGA        & 0.6881 & 0.5867 & 0.5148 & 0.4845 & 0.3962 & 0.3979 \\
  &               & HANAttack  & 0.5000 & 0.3460 & 0.7537 & 0.7505 & 0.5151 & 0.4683 \\
  &               & HeteroRL(w/o Top-K)   & $0.3688^\dagger$ & $0.2889^\dagger$ & $0.2931^\dagger$ & $0.2099^\dagger$ & $0.3473^\dagger$ & $0.3395^\dagger$ \\
  &               & HeteroKRL  & \textbf{0.0965} & \textbf{0.0667} & \textbf{0.0271} & \textbf{0.0243} & \textbf{0.0070} & \textbf{0.0072} \\
\cline{2-9}
  & \multirow{5}{*}{5}
    & Random     & 0.7623 & 0.7394 & 0.4950 & 0.4777 & 0.5804 & 0.5765 \\
  &               & FGA        & 0.5074 & 0.2550 & 0.4384 & 0.4078 & $0.2937\dagger$ & $0.2875\dagger$ \\
  &               & HANAttack  & 0.5173 & 0.2948 & 0.4187 & 0.3775 & 0.5128 & 0.4783 \\
  &               & HeteroRL(w/o Top-K)   & $0.1436^\dagger$ & $0.1021^\dagger$ & $0.2758^\dagger$ & $0.2194^\dagger$ & 0.3450 & 0.3327 \\
  &               & HeteroKRL  & \textbf{0.1089} & \textbf{0.0726} & \textbf{0.0246} & \textbf{0.0216} & \textbf{0.0280} & \textbf{0.0263} \\
\cline{1-9}
\end{tabular}
\caption{Attack results on three datasets (ACM, DBLP, IMDB).}
\label{tab:attack_results}
\end{table*}

%%%%%%%%%%%%%%%%%%%%%%%%%%%%%%%%%%%%%%%%%%%%%%%%%%%%%%%%%%%%%%%%

\section{Experiments}
In this section, we evaluate our proposed method across multiple heterogeneous graph benchmarks to investigate the following research questions:

\begin{enumerate}[label=\textbf{RQ\arabic*:}, leftmargin=0.5cm]
    \item How effective is the proposed attack in degrading model performance on target nodes?
    \item What is the impact of removing the Top-K refinement module from the framework?
    \item How does the attack effectiveness vary with respect to the hyperparameter $K$?
    \item How resilient are existing heterogeneous graph defense methods against the proposed attack?
    \item Can an attack policy trained on one model transfer effectively to other models?
    \item What selection preferences does the RL agent exhibit during edge perturbation?
\end{enumerate}

\subsection{Experiment Settings}
We conduct experiments on three benchmark heterogeneous graph datasets: ACM, IMDB, and DBLP. ACM contains papers categorized into Database, Wireless Communication, and Data Mining. IMDB consists of movies labeled by genre (Action, Comedy, Drama), along with related entities such as actors and directors. DBLP includes papers classified into Database, Artificial Intelligence, and Information Retrieval. Detailed dataset statistics are summarized in Table~\ref{tab:dataset statistics}. We evaluate our method against three representative victim models: HAN~\citep{wang2019heterogeneous}, HGT~\citep{heterogeneousgraphtransformer}, and SimpleHGN~\citep{SimpleHGN}, covering a range of attention mechanisms and relation modeling techniques. All experiments are conducted on a workstation equipped with an AMD Ryzen Threadripper PRO 3955WX (16 cores), 32GB RAM, and an NVIDIA RTX 3090 GPU (24GB VRAM). The detailed model-specific training configurations are provided in Appendix~\ref{appendix: experiment_victim_settings}.

\begin{table}[t]
\centering
\small
\resizebox{\linewidth}{!}{ % 强制调整表格宽度
\begin{tabular}{lccccc}
\toprule
\textbf{Dataset} & \textbf{\#Node Types} & \textbf{\#Edge Types} & \textbf{\#Nodes} & \textbf{\#Edges} & \textbf{Primary Type} \\
\midrule
ACM   & 3 & 4 & 11252  & 34864  & paper  \\
IMDB  & 3 & 4 & 11616  & 34212  & movie \\
DBLP  & 4 & 6 & 26198  & 242142 & author  \\
\bottomrule
\end{tabular}
}
\caption{Dataset Statistics}
\label{tab:dataset statistics}
\end{table}

\subsection{Attack Performance}
\begin{figure*}[t]
    \centering
    \includegraphics[width=0.32\textwidth]{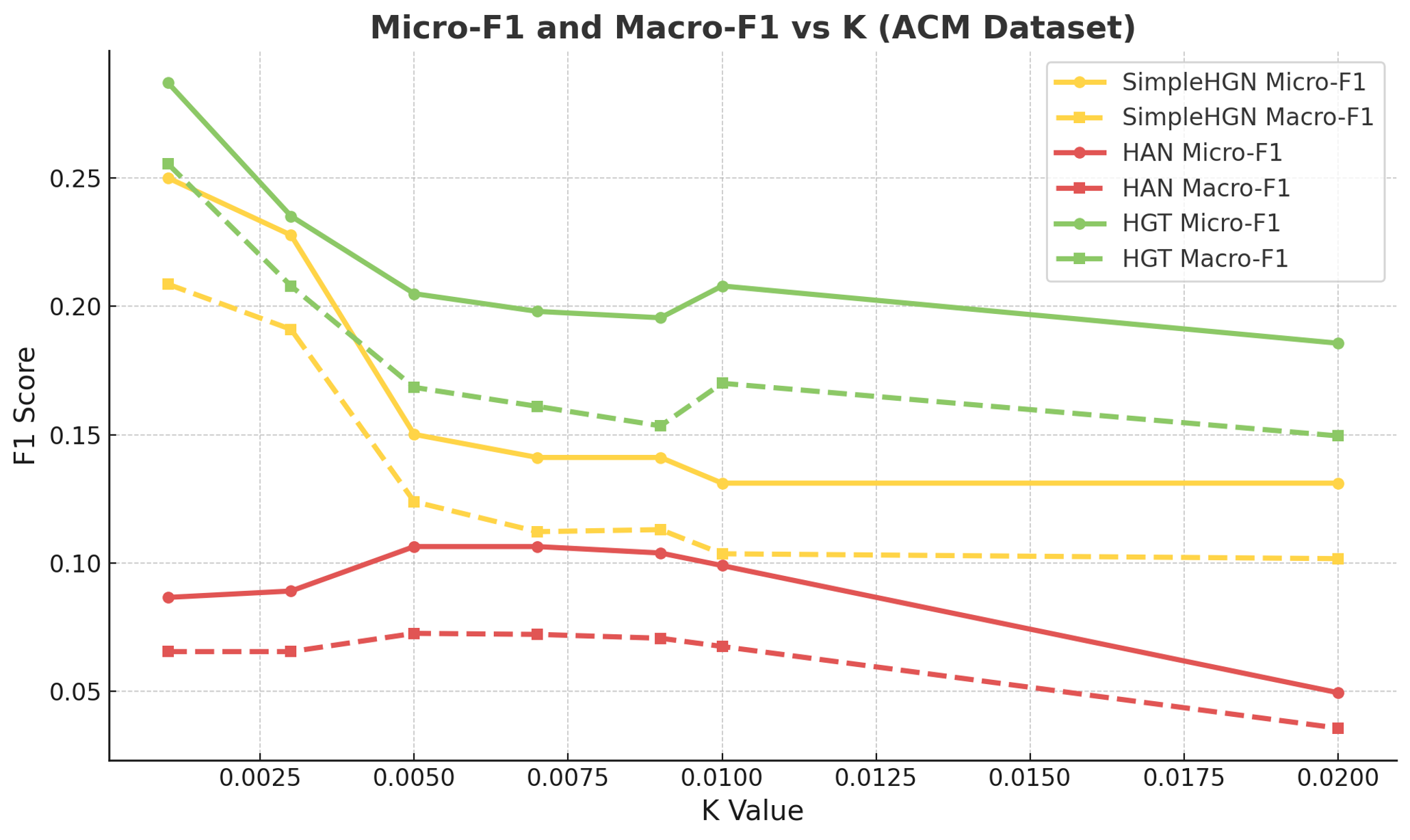}
    \includegraphics[width=0.32\textwidth]{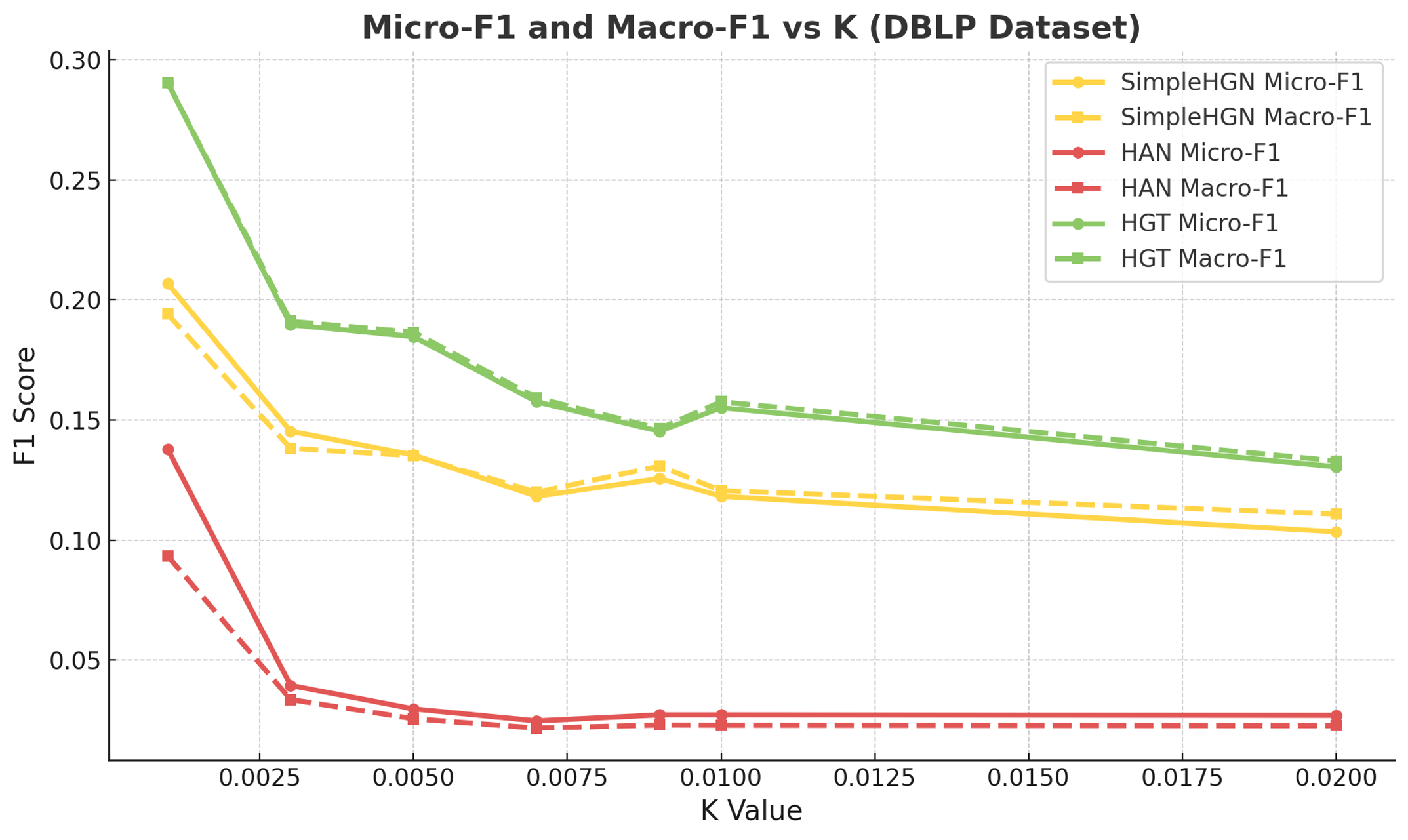}
    \includegraphics[width=0.32\textwidth]{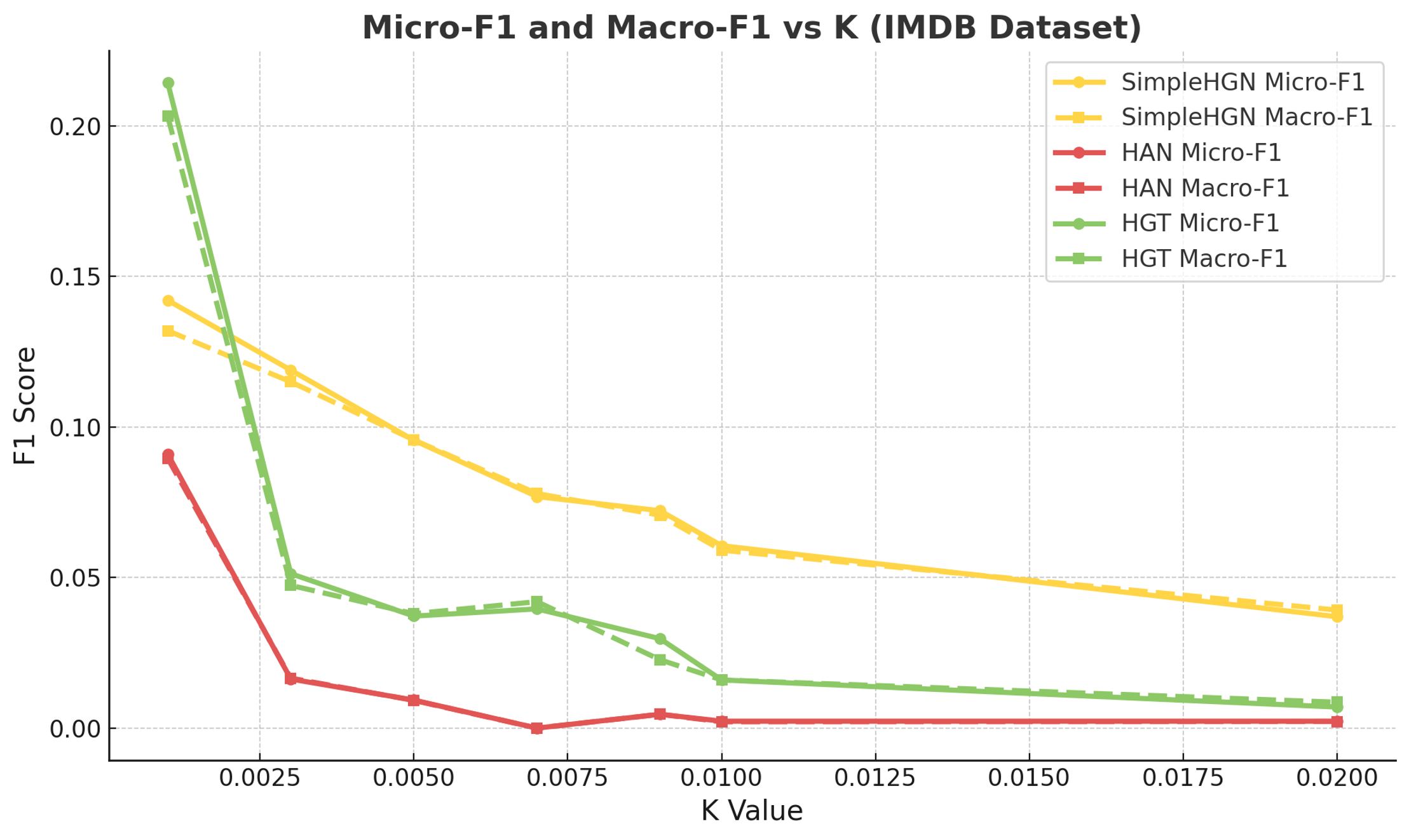}
    \caption{Comparison of models with different K values for IMDB, ACM, and DBLP datasets}
    \label{fig:ablation comparison}
\end{figure*}
To answer the RQ1, We use the Micro-F1 and Macro-F1 scores of the target classification model (i.e., the victim model) as the primary evaluation metrics. Micro-F1 measures overall prediction accuracy, while Macro-F1 captures the average performance across all classes. Together, these metrics reflect the degree to which an attack degrades the classification capability of the model.

To evaluate the effectiveness of our proposed method, HeteroKRL, we compare it against two representative gray-box attack baselines. The first baseline, FGA (Fast Gradient Attack) ~\citep{FGA}, perturbs the graph structure by computing gradients of the adjacency matrix with respect to the model's loss function, identifying and modifying the most influential edges. Since FGA is originally designed for homogeneous graphs, we convert the heterogeneous graph into a homogeneous format before applying this attack. The second baseline is HGAttack~\citep{zhao2024hgattack}, a gray-box attack method tailored for heterogeneous graphs. It employs a surrogate model and uses metapath-based gradient analysis to locate critical edges for perturbation.

As shown in Table~\ref{tab:attack_results}, we bold the lowest performance values (i.e., the most successful attacks) under each setting to emphasize comparative attack efficacy. Despite operating under more restricted conditions than gray-box attacks—namely, without access to the complete heterogeneous graph structure (i.e., adjacency information)—our black-box method HeteroKRL consistently outperforms both baselines across all datasets (ACM, DBLP, IMDB), all victim models (SimpleHGN, HGT, HAN), and various perturbation budgets. HeteroKRL achieves greater degradation in both Micro-F1 and Macro-F1, clearly demonstrating its strong attack capability even under limited access scenarios.

\subsection{Ablation Study}
For RQ2, We conduct an ablation study to evaluate the effectiveness of the Top-K strategy used in our proposed method. As shown in Table~\ref{tab:attack_results}, the best results among methods other than HeteroKRL are marked with $\dagger$. Although HeteroRL (w/o Top-K) already outperforms both gray-box baselines under most settings, incorporating the Top-K mechanism significantly enhances the attack performance. For instance, when the victim model is HAN, the dataset is ACM, and the budget is 3, the Micro-F1 of HeteroRL reaches 0.3688, while HeteroKRL with Top-K achieves a much lower Micro-F1 of only 0.0965.

This performance gain can be attributed to the fact that the reinforcement learning agent has learned which node features are most influential to the victim node’s prediction. By leveraging the Top-K strategy to guide the search toward nodes with similar impactful features, the agent can more efficiently identify disruptive structural modifications, thereby improving the overall attack effectiveness.

\subsection{Hyperparameter Analysis}

In this section, we investigate how the hyperparameter $K$ affects the attack performance to answer RQ3. Here, $K$ denotes the proportion of candidate nodes selected from each auxiliary type. For example, if $K = 0.001$ and there are 3000 nodes of a specific auxiliary type, we select $K \cdot |\mathcal{V}_{t_{\text{aux}}}| = 3$ nodes as candidate nodes for potential edge modification.

We evaluate the impact of $K$ by varying its value from 0.001 to 0.02 across the three datasets: ACM, DBLP, and IMDB. As shown in Figure~\ref{fig:ablation comparison}, increasing $K$ leads to a general decrease in both Micro-F1 and Macro-F1 scores, indicating stronger attack performance. However, in nearly all cases, we observe that the performance drop becomes marginal once $K$ exceeds a certain threshold (e.g., 0.003 or 0.005), showing clear diminishing returns.

For example, on the DBLP dataset with HAN as the victim model, increasing $K$ from 0.001 to 0.003 results in a significant drop in F1 scores, reflecting a strong gain in attack effectiveness. In contrast, further increasing $K$ beyond this point yields only minor improvements. This observation supports our earlier hypothesis: the RL agent has learned which node features are most influential to the victim node. When $K$ is small, candidate nodes are more likely to be close to those influential features, resulting in efficient and effective attacks. As $K$ increases, the search space becomes larger and more diluted, including nodes with less relevant features, which reduces the marginal gains in attack performance.

\subsection{Attack on Defense Strategies}
\begin{table}[ht]
\centering
\scriptsize
\begin{tabular}{clcccc}
\hline
\multirow{2}{*}{\textbf{Victim Model}} 
  & \multirow{2}{*}{\textbf{Attack Method}} 
  & \multicolumn{2}{c}{\textbf{ACM}}
  & \multicolumn{2}{c}{\textbf{IMDB}} \\ 
\cline{3-6}
  & & Micro-F1 & Macro-F1 & Micro-F1 & Macro-F1 \\ 
\hline
\multirow{6}{*}{RoHE}
  & Clean      & 0.8985 & 0.9015 & 0.6223 & 0.6224 \\
  & Random     & 0.8811 & 0.8818 & 0.5967 & 0.5955 \\
  & FGA        & 0.8787 & 0.8799 & 0.4615 & 0.4654 \\
  & HANAttack  & 0.8762 & 0.8786 & 0.5944 & 0.5931 \\
  & HeteroRL   & 0.4455 & 0.2679 & 0.4988 & 0.5015 \\
  & HeteroKRL  & \textbf{0.3861} & \textbf{0.2413} & \textbf{0.1814} & \textbf{0.1912} \\
\hline
\end{tabular}
\caption{Attack results of RoHe on ACM and IMDB datasets.}
\label{tab:RoHE_results}
\end{table}

To address RQ4, we apply our proposed attack method to RoHe \citep{zhang2022robust}, a defense framework specifically designed to enhance the robustness of HGNNs against topological adversarial attacks. RoHe mitigates adversarial influence by incorporating an attention purifier, which prunes potentially malicious neighbors based on both topological structure and feature similarity.

In our experiments, we set the attack budget to 5 and evaluate on the ACM and IMDB datasets. The DBLP dataset is excluded due to its long training time under the RoHe framework.

As shown in Table~\ref{tab:RoHE_results}, RoHe demonstrates strong resistance to baseline attacks such as FGA and HANAttack, maintaining high Micro-F1 and Macro-F1 scores. However, it is significantly less effective against our proposed HeteroKRL method. For example, on the ACM dataset, the Micro-F1 score drops sharply from nearly 90\% to 38\%, revealing a substantial vulnerability. In contrast, the performance drop under FGA and HANAttack is minor, suggesting that existing defenses are insufficient against our attack strategy.

\subsection{Transferability}
\begin{table}[htbp]
  \centering
  \scriptsize
  \setlength{\tabcolsep}{2.5pt}     % 缩小列间距
  \renewcommand{\arraystretch}{1.1} % 压缩行高
  % 强制缩放到当前栏宽度
  \resizebox{\columnwidth}{!}{%
    \begin{tabular}{ll cc cc cc}
      \toprule
      \multirow{2}{*}{\textbf{Transfer Model}} & \multirow{2}{*}{\textbf{Target Model}}
        & \multicolumn{2}{c}{\textbf{ACM}}
        & \multicolumn{2}{c}{\textbf{DBLP}}
        & \multicolumn{2}{c}{\textbf{IMDB}} \\
      \cline{3-8}
      & & Micro-F1 & Macro-F1 & Micro-F1 & Macro-F1 & Micro-F1 & Macro-F1 \\
      \midrule
      \multirow{2}{*}{SimpleHGN}
        & HAN        & 0.2673 & 0.1626 & 0.0295 & 0.0249 & 0.0093 & 0.0088 \\
        & HGT        & 0.2797 & 0.2072 & 0.1798 & 0.1802 & 0.0466 & 0.0496 \\
      \midrule
      \multirow{2}{*}{HAN}
        & SimpleHGN  & 0.1039 & 0.0774 & 0.1355 & 0.1358 & 0.1585 & 0.1590 \\
        & HGT        & 0.2029 & 0.1417 & 0.1724 & 0.1784 & 0.0419 & 0.0447 \\
      \midrule
      \multirow{2}{*}{HGT}
        & HAN        & 0.0346 & 0.0302 & 0.0246 & 0.0208 & 0.1818 & 0.1808 \\
        & SimpleHGN  & 0.1064 & 0.0855 & 0.1502 & 0.1523 & 0.0186 & 0.0165 \\
      \bottomrule
    \end{tabular}%
  }
  \caption{Transfer performance (Micro‐F1, Macro‐F1) of models across ACM, DBLP and IMDB datasets.}
  \label{tab:transfer-performance}
\end{table}
To address RQ5, we investigate the transferability of our proposed HeteroRL attack agent across different heterogeneous graph neural network (HGNN) models. Specifically, we train the attack agent on a given transfer model and evaluate its effectiveness on a different target model, without any further retraining or adaptation. During testing, we set the attack budget to 3 and employ a Top-K selection strategy to identify the most impactful perturbations.

As shown in Table~\ref{tab:transfer-performance}, the results demonstrate clear cross-model transferability. For instance, an agent trained on SimpleHGN can significantly degrade the performance of HAN and HGT, and vice versa. Consistent drops in Micro-F1 and Macro-F1 are observed across ACM, DBLP, and IMDB datasets.

This indicates that simply switching model architectures is insufficient for defense. We attribute this transferability to the shared use of attention mechanisms in HAN, HGT, and SimpleHGN. Although their aggregation strategies differ (e.g., metapath-level in HAN and type-level in HGT), they share structural similarities that allow attack strategies to generalize across models.

The strong transferability of our method highlights a critical weakness in current HGNN architectures: structural similarity in their reliance on attention introduces a common attack surface, which can be leveraged even without white-box knowledge of the target model. This reinforces the need for defense mechanisms that are not only model-specific but also capable of addressing shared vulnerabilities across attention-based models.

\subsection{Data Analysis}
\begin{figure}[ht]
  \centering
  \begin{minipage}[b]{0.48\columnwidth}
    \centering
    \includegraphics[width=\linewidth]{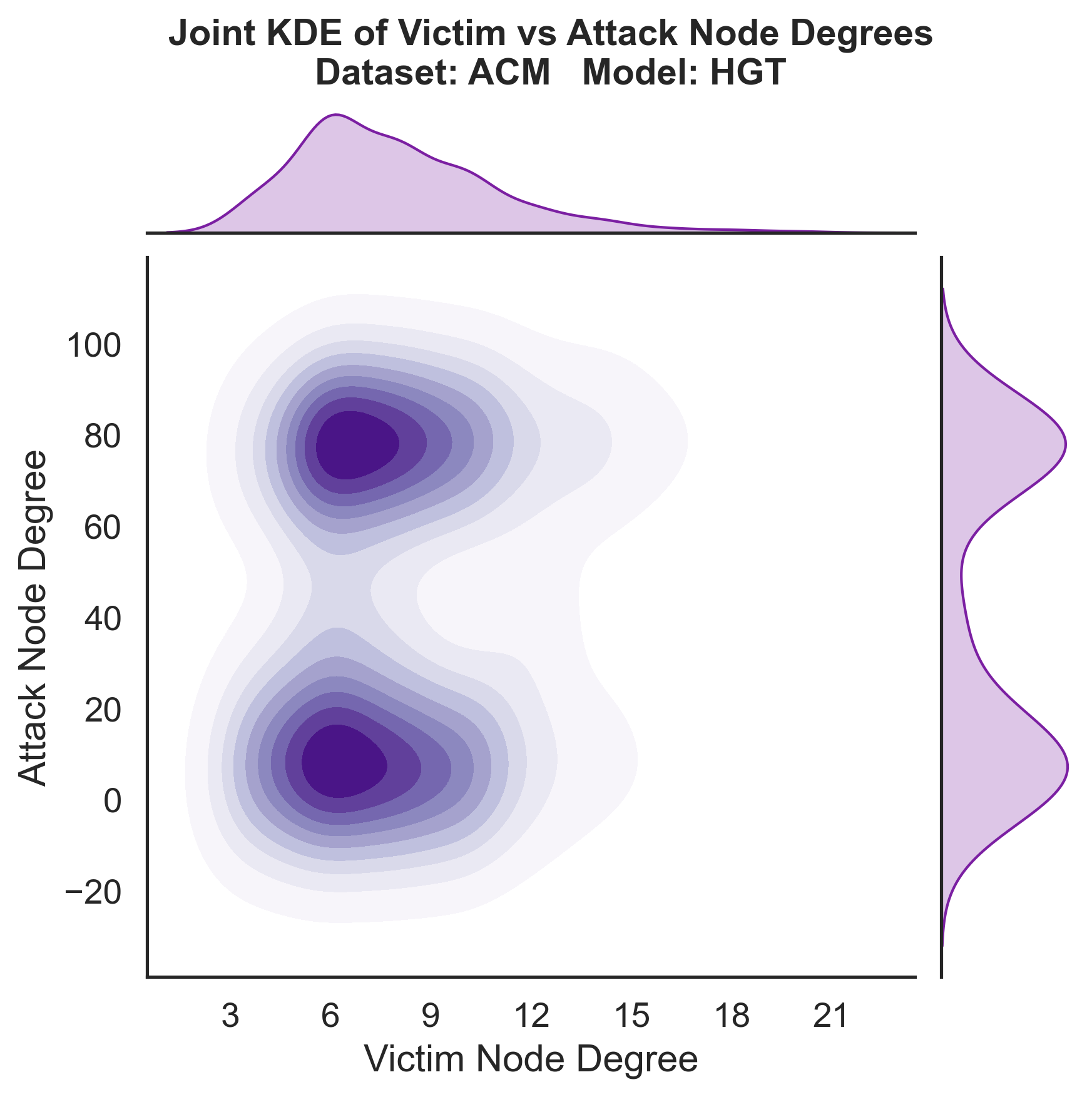}
    \label{fig:sub1}
  \end{minipage}%
  \hfill
  \begin{minipage}[b]{0.48\columnwidth}
    \centering
    \includegraphics[width=\linewidth]{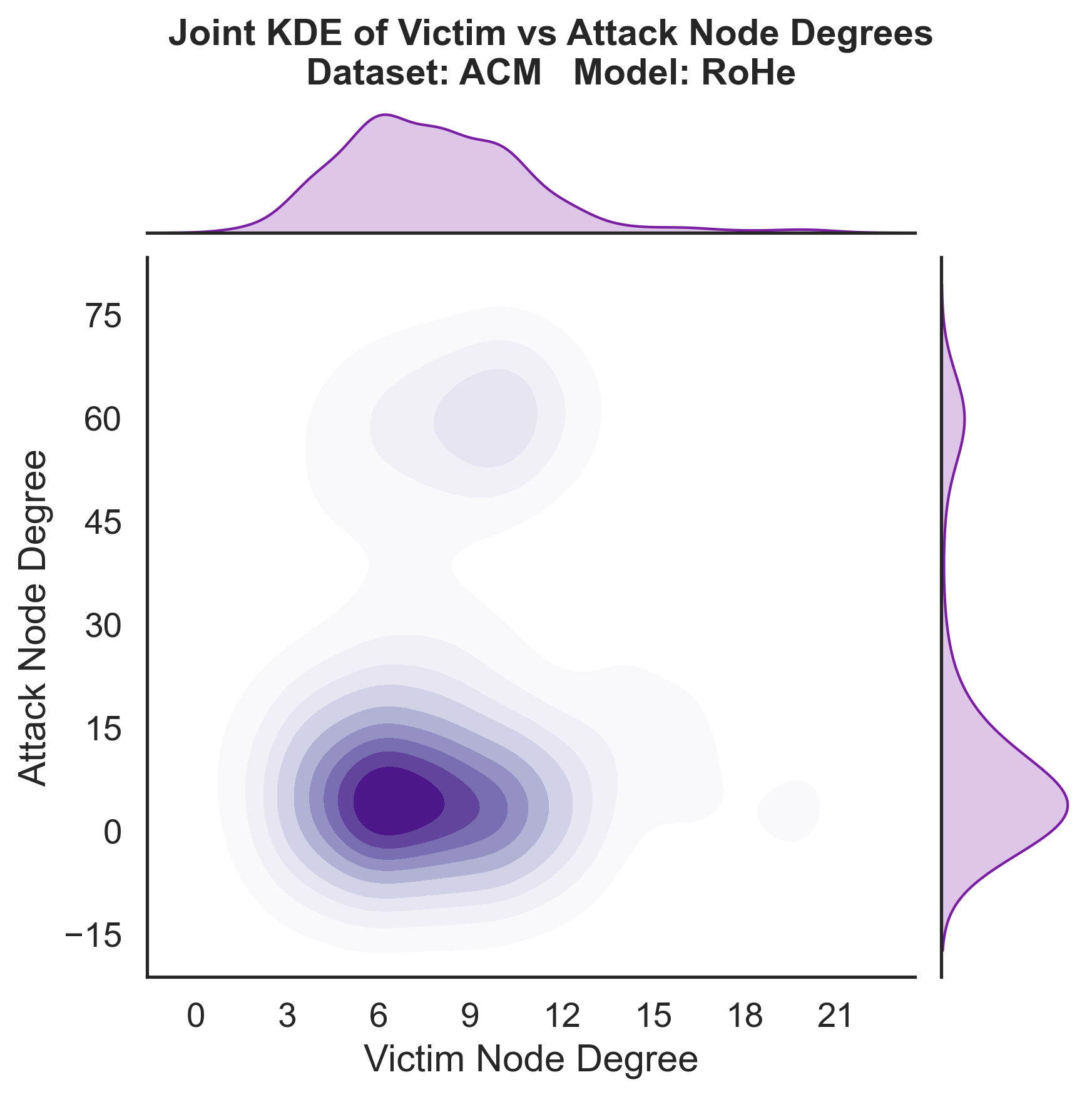}
    \label{fig:sub2}
  \end{minipage}
  \caption{Joint KDE visualization on the ACM dataset for two attack methods.}
  \label{fig:kde-ana}
\end{figure}
To address RQ6, we analyze the joint degree distribution of successfully attacked target nodes and their corresponding attacker nodes on the ACM dataset in Figure \ref{fig:kde-ana}. From the results on both HGT and RoHe, we observe that adversarial paths often involve high-degree nodes (i.e., hubs), as such nodes exert stronger influence during message aggregation \citep{degreeAndInfluence}.

However, we also identify another important pattern: some low-degree nodes can still possess considerable influence. This is because such nodes may act as bridging points in the graph, connecting to multiple high-impact neighbors and playing a critical intermediary role in feature propagation. In heterogeneous graphs, the interplay between structural and semantic factors further amplifies the attention these nodes receive, making them highly influential despite their low degree.

As a result, although RoHe adopts a degree-based purification strategy, it may still fail to detect these "low-degree but high-impact" nodes, allowing adversarial interference to persist. This observation further supports the effectiveness of our attack method against defense-enhanced models like RoHe.

\section{Conclusion}
We propose HeteroKRLAttack, a reinforcement learning-based black-box evasion attack tailored for heterogeneous graphs. Without requiring access to model parameters or complete graph structure, our method effectively identifies influential perturbations via a hierarchical policy network and a Top-K refinement strategy. Experiments across three benchmark datasets and multiple victim models show that HeteroKRLAttack consistently outperforms existing gray-box attacks, and remains effective even against defense-enhanced models like RoHe.

Our work highlights critical vulnerabilities in current HGNNs under realistic attack scenarios and provides a generalizable attack strategy with strong transferability. This raises security concerns for HGNN applications in sensitive domains such as finance, healthcare, and recommendation systems. Future work will explore more robust, model-agnostic defenses and extend our method to dynamic and graph-level attack tasks.

\newpage
\bibliographystyle{ACM-Reference-Format}
\bibliography{sample-base}


\begin{thebibliography}{44}

%%% the Web address.

\ifx \showCODEN    \undefined \def \showCODEN     #1{\unskip}     \fi
\ifx \showDOI      \undefined \def \showDOI       #1{#1}\fi
\ifx \showISBNx    \undefined \def \showISBNx     #1{\unskip}     \fi
\ifx \showISBNxiii \undefined \def \showISBNxiii  #1{\unskip}     \fi
\ifx \showISSN     \undefined \def \showISSN      #1{\unskip}     \fi
\ifx \showLCCN     \undefined \def \showLCCN      #1{\unskip}     \fi
\ifx \shownote     \undefined \def \shownote      #1{#1}          \fi
\ifx \showarticletitle \undefined \def \showarticletitle #1{#1}   \fi
\ifx \showURL      \undefined \def \showURL       {\relax}        \fi
% invisible to TeX
\providecommand\bibfield[2]{#2}
\providecommand\bibinfo[2]{#2}
\providecommand\natexlab[1]{#1}
\providecommand\showeprint[2][]{arXiv:#2}

\bibitem[Awasthi et~al\mbox{.}(2023)]%
        {socialNet3}
\bibfield{author}{\bibinfo{person}{AK Awasthi}, \bibinfo{person}{Arun~Kumar Garov}, \bibinfo{person}{Minakshi Sharma}, {and} \bibinfo{person}{Mrigank Sinha}.} \bibinfo{year}{2023}\natexlab{}.
\newblock \showarticletitle{GNN model based on node classification forecasting in social network}. In \bibinfo{booktitle}{\emph{2023 International Conference on Artificial Intelligence and Smart Communication (AISC)}}. IEEE, \bibinfo{pages}{1039--1043}.
\newblock


\bibitem[Bamakan et~al\mbox{.}(2019)]%
        {degreeAndInfluence}
\bibfield{author}{\bibinfo{person}{Seyed Mojtaba~Hosseini Bamakan}, \bibinfo{person}{Ildar Nurgaliev}, {and} \bibinfo{person}{Qiang Qu}.} \bibinfo{year}{2019}\natexlab{}.
\newblock \showarticletitle{Opinion leader detection: A methodological review}.
\newblock \bibinfo{journal}{\emph{Expert Systems with Applications}}  \bibinfo{volume}{115} (\bibinfo{year}{2019}), \bibinfo{pages}{200--222}.
\newblock


\bibitem[Baniecki and Biecek(2024)]%
        {adversarialAttackDefinition}
\bibfield{author}{\bibinfo{person}{Hubert Baniecki} {and} \bibinfo{person}{Przemyslaw Biecek}.} \bibinfo{year}{2024}\natexlab{}.
\newblock \showarticletitle{Adversarial attacks and defenses in explainable artificial intelligence: A survey}.
\newblock \bibinfo{journal}{\emph{Information Fusion}} (\bibinfo{year}{2024}), \bibinfo{pages}{102303}.
\newblock


\bibitem[Bhagat et~al\mbox{.}(2011)]%
        {NodeclassificationinSocialNetworks}
\bibfield{author}{\bibinfo{person}{Smriti Bhagat}, \bibinfo{person}{Graham Cormode}, {and} \bibinfo{person}{S Muthukrishnan}.} \bibinfo{year}{2011}\natexlab{}.
\newblock \showarticletitle{Node classification in social networks}.
\newblock \bibinfo{journal}{\emph{Social network data analytics}} (\bibinfo{year}{2011}), \bibinfo{pages}{115--148}.
\newblock


\bibitem[Bongini et~al\mbox{.}(2022)]%
        {biognn2}
\bibfield{author}{\bibinfo{person}{Pietro Bongini}, \bibinfo{person}{Niccol{\`o} Pancino}, \bibinfo{person}{Franco Scarselli}, {and} \bibinfo{person}{Monica Bianchini}.} \bibinfo{year}{2022}\natexlab{}.
\newblock \showarticletitle{BioGNN: how graph neural networks can solve biological problems}.
\newblock In \bibinfo{booktitle}{\emph{Artificial Intelligence and Machine Learning for Healthcare: Vol. 1: Image and Data Analytics}}. \bibinfo{publisher}{Springer}, \bibinfo{pages}{211--231}.
\newblock


\bibitem[Chen et~al\mbox{.}(2018)]%
        {FGA}
\bibfield{author}{\bibinfo{person}{Jinyin Chen}, \bibinfo{person}{Yangyang Wu}, \bibinfo{person}{Xuanheng Xu}, \bibinfo{person}{Yixian Chen}, \bibinfo{person}{Haibin Zheng}, {and} \bibinfo{person}{Qi Xuan}.} \bibinfo{year}{2018}\natexlab{}.
\newblock \showarticletitle{Fast gradient attack on network embedding}.
\newblock \bibinfo{journal}{\emph{arXiv preprint arXiv:1809.02797}} (\bibinfo{year}{2018}).
\newblock


\bibitem[Dai et~al\mbox{.}(2018)]%
        {dai2018adversarial}
\bibfield{author}{\bibinfo{person}{Hanjun Dai}, \bibinfo{person}{Hui Li}, \bibinfo{person}{Tian Tian}, \bibinfo{person}{Xin Huang}, \bibinfo{person}{Lin Wang}, \bibinfo{person}{Jun Zhu}, {and} \bibinfo{person}{Le Song}.} \bibinfo{year}{2018}\natexlab{}.
\newblock \showarticletitle{Adversarial attack on graph structured data}. In \bibinfo{booktitle}{\emph{International conference on machine learning}}. PMLR, \bibinfo{pages}{1115--1124}.
\newblock


\bibitem[Dulac-Arnold et~al\mbox{.}(2015)]%
        {2015deepRL}
\bibfield{author}{\bibinfo{person}{Gabriel Dulac-Arnold}, \bibinfo{person}{Richard Evans}, \bibinfo{person}{Hado van Hasselt}, \bibinfo{person}{Peter Sunehag}, \bibinfo{person}{Timothy Lillicrap}, \bibinfo{person}{Jonathan Hunt}, \bibinfo{person}{Timothy Mann}, \bibinfo{person}{Theophane Weber}, \bibinfo{person}{Thomas Degris}, {and} \bibinfo{person}{Ben Coppin}.} \bibinfo{year}{2015}\natexlab{}.
\newblock \showarticletitle{Deep reinforcement learning in large discrete action spaces}.
\newblock \bibinfo{journal}{\emph{arXiv preprint arXiv:1512.07679}} (\bibinfo{year}{2015}).
\newblock


\bibitem[Fan et~al\mbox{.}(2019)]%
        {gnnandsocial}
\bibfield{author}{\bibinfo{person}{Wenqi Fan}, \bibinfo{person}{Yao Ma}, \bibinfo{person}{Qing Li}, \bibinfo{person}{Yuan He}, \bibinfo{person}{Eric Zhao}, \bibinfo{person}{Jiliang Tang}, {and} \bibinfo{person}{Dawei Yin}.} \bibinfo{year}{2019}\natexlab{}.
\newblock \showarticletitle{Graph neural networks for social recommendation}. In \bibinfo{booktitle}{\emph{The world wide web conference}}. \bibinfo{pages}{417--426}.
\newblock


\bibitem[Friedman et~al\mbox{.}(1976)]%
        {kdtree}
\bibfield{author}{\bibinfo{person}{Jerome~H Friedman}, \bibinfo{person}{Jon~Louis Bentley}, {and} \bibinfo{person}{Raphael~Ari Finkel}.} \bibinfo{year}{1976}\natexlab{}.
\newblock \showarticletitle{An algorithm for finding best matches in logarithmic time}.
\newblock \bibinfo{journal}{\emph{ACM Trans. Math. Software}} \bibinfo{volume}{3}, \bibinfo{number}{SLAC-PUB-1549-REV. 2} (\bibinfo{year}{1976}), \bibinfo{pages}{209--226}.
\newblock


\bibitem[Fu et~al\mbox{.}(2020)]%
        {MAGNN}
\bibfield{author}{\bibinfo{person}{Xinyu Fu}, \bibinfo{person}{Jiani Zhang}, \bibinfo{person}{Ziqiao Meng}, {and} \bibinfo{person}{Irwin King}.} \bibinfo{year}{2020}\natexlab{}.
\newblock \showarticletitle{Magnn: Metapath aggregated graph neural network for heterogeneous graph embedding}. In \bibinfo{booktitle}{\emph{Proceedings of the web conference 2020}}. \bibinfo{pages}{2331--2341}.
\newblock


\bibitem[Geng et~al\mbox{.}(2022)]%
        {HGNNandCite}
\bibfield{author}{\bibinfo{person}{Hao Geng}, \bibinfo{person}{Deqing Wang}, \bibinfo{person}{Fuzhen Zhuang}, \bibinfo{person}{Xuehua Ming}, \bibinfo{person}{Chenguang Du}, \bibinfo{person}{Ting Jiang}, \bibinfo{person}{Haolong Guo}, {and} \bibinfo{person}{Rui Liu}.} \bibinfo{year}{2022}\natexlab{}.
\newblock \showarticletitle{Modeling dynamic heterogeneous graph and node importance for future citation prediction}. In \bibinfo{booktitle}{\emph{Proceedings of the 31st ACM International Conference on Information \& Knowledge Management}}. \bibinfo{pages}{572--581}.
\newblock


\bibitem[Goodfellow et~al\mbox{.}(2016)]%
        {Goodfellow-et-al-2016}
\bibfield{author}{\bibinfo{person}{Ian Goodfellow}, \bibinfo{person}{Yoshua Bengio}, {and} \bibinfo{person}{Aaron Courville}.} \bibinfo{year}{2016}\natexlab{}.
\newblock \bibinfo{booktitle}{\emph{Deep Learning}}.
\newblock \bibinfo{publisher}{MIT Press}.
\newblock
\showISBNx{9780262035613}
\urldef\tempurl%
\url{http://www.deeplearningbook.org}
\showURL{%
\tempurl}


\bibitem[Goodfellow et~al\mbox{.}(2014)]%
        {Goodfellow2014ExplainingAH}
\bibfield{author}{\bibinfo{person}{Ian~J. Goodfellow}, \bibinfo{person}{Jonathon Shlens}, {and} \bibinfo{person}{Christian Szegedy}.} \bibinfo{year}{2014}\natexlab{}.
\newblock \showarticletitle{Explaining and Harnessing Adversarial Examples}. In \bibinfo{booktitle}{\emph{International Conference on Learning Representations (ICLR)}}.
\newblock


\bibitem[Hu et~al\mbox{.}(2024)]%
        {RpHGNN}
\bibfield{author}{\bibinfo{person}{Jun Hu}, \bibinfo{person}{Bryan Hooi}, {and} \bibinfo{person}{Bingsheng He}.} \bibinfo{year}{2024}\natexlab{}.
\newblock \showarticletitle{Efficient heterogeneous graph learning via random projection}.
\newblock \bibinfo{journal}{\emph{IEEE Transactions on Knowledge and Data Engineering}} (\bibinfo{year}{2024}).
\newblock


\bibitem[Hu et~al\mbox{.}(2020)]%
        {heterogeneousgraphtransformer}
\bibfield{author}{\bibinfo{person}{Ziniu Hu}, \bibinfo{person}{Yuxiao Dong}, \bibinfo{person}{Kuansan Wang}, {and} \bibinfo{person}{Yizhou Sun}.} \bibinfo{year}{2020}\natexlab{}.
\newblock \showarticletitle{Heterogeneous graph transformer}. In \bibinfo{booktitle}{\emph{Proceedings of the web conference 2020}}. \bibinfo{pages}{2704--2710}.
\newblock


\bibitem[Li and Pi(2019)]%
        {userinterest1}
\bibfield{author}{\bibinfo{person}{Bentian Li} {and} \bibinfo{person}{Dechang Pi}.} \bibinfo{year}{2019}\natexlab{}.
\newblock \showarticletitle{Learning deep neural networks for node classification}.
\newblock \bibinfo{journal}{\emph{Expert Systems with Applications}}  \bibinfo{volume}{137} (\bibinfo{year}{2019}), \bibinfo{pages}{324--334}.
\newblock


\bibitem[Li et~al\mbox{.}(2021)]%
        {biognn3}
\bibfield{author}{\bibinfo{person}{Rui Li}, \bibinfo{person}{Xin Yuan}, \bibinfo{person}{Mohsen Radfar}, \bibinfo{person}{Peter Marendy}, \bibinfo{person}{Wei Ni}, \bibinfo{person}{Terrence~J O’Brien}, {and} \bibinfo{person}{Pablo~M Casillas-Espinosa}.} \bibinfo{year}{2021}\natexlab{}.
\newblock \showarticletitle{Graph signal processing, graph neural network and graph learning on biological data: a systematic review}.
\newblock \bibinfo{journal}{\emph{IEEE Reviews in Biomedical Engineering}}  \bibinfo{volume}{16} (\bibinfo{year}{2021}), \bibinfo{pages}{109--135}.
\newblock


\bibitem[Liu et~al\mbox{.}(2021)]%
        {knowledgegraphgnn2}
\bibfield{author}{\bibinfo{person}{Shuwen Liu}, \bibinfo{person}{Bernardo Grau}, \bibinfo{person}{Ian Horrocks}, {and} \bibinfo{person}{Egor Kostylev}.} \bibinfo{year}{2021}\natexlab{}.
\newblock \showarticletitle{Indigo: Gnn-based inductive knowledge graph completion using pair-wise encoding}.
\newblock \bibinfo{journal}{\emph{Advances in Neural Information Processing Systems}}  \bibinfo{volume}{34} (\bibinfo{year}{2021}), \bibinfo{pages}{2034--2045}.
\newblock


\bibitem[Lv et~al\mbox{.}(2021)]%
        {SimpleHGN}
\bibfield{author}{\bibinfo{person}{Qingsong Lv}, \bibinfo{person}{Ming Ding}, \bibinfo{person}{Qiang Liu}, \bibinfo{person}{Yuxiang Chen}, \bibinfo{person}{Wenzheng Feng}, \bibinfo{person}{Siming He}, \bibinfo{person}{Chang Zhou}, \bibinfo{person}{Jianguo Jiang}, \bibinfo{person}{Yuxiao Dong}, {and} \bibinfo{person}{Jie Tang}.} \bibinfo{year}{2021}\natexlab{}.
\newblock \showarticletitle{Are we really making much progress? revisiting, benchmarking and refining heterogeneous graph neural networks}. In \bibinfo{booktitle}{\emph{Proceedings of the 27th ACM SIGKDD conference on knowledge discovery \& data mining}}. \bibinfo{pages}{1150--1160}.
\newblock


\bibitem[Ma et~al\mbox{.}(2021)]%
        {ma2021graph}
\bibfield{author}{\bibinfo{person}{Yao Ma}, \bibinfo{person}{Suhang Wang}, \bibinfo{person}{Tyler Derr}, \bibinfo{person}{Lingfei Wu}, {and} \bibinfo{person}{Jiliang Tang}.} \bibinfo{year}{2021}\natexlab{}.
\newblock \showarticletitle{Graph adversarial attack via rewiring}. In \bibinfo{booktitle}{\emph{Proceedings of the 27th ACM SIGKDD Conference on Knowledge Discovery \& Data Mining}}. \bibinfo{pages}{1161--1169}.
\newblock


\bibitem[Muzio et~al\mbox{.}(2021a)]%
        {gnnandbiological}
\bibfield{author}{\bibinfo{person}{Giulia Muzio}, \bibinfo{person}{Leslie O’Bray}, {and} \bibinfo{person}{Karsten Borgwardt}.} \bibinfo{year}{2021}\natexlab{a}.
\newblock \showarticletitle{Biological network analysis with deep learning}.
\newblock \bibinfo{journal}{\emph{Briefings in bioinformatics}} \bibinfo{volume}{22}, \bibinfo{number}{2} (\bibinfo{year}{2021}), \bibinfo{pages}{1515--1530}.
\newblock


\bibitem[Muzio et~al\mbox{.}(2021b)]%
        {NodeClassificationAndBiologicalNet}
\bibfield{author}{\bibinfo{person}{Giulia Muzio}, \bibinfo{person}{Leslie O’Bray}, {and} \bibinfo{person}{Karsten Borgwardt}.} \bibinfo{year}{2021}\natexlab{b}.
\newblock \showarticletitle{Biological network analysis with deep learning}.
\newblock \bibinfo{journal}{\emph{Briefings in bioinformatics}} \bibinfo{volume}{22}, \bibinfo{number}{2} (\bibinfo{year}{2021}), \bibinfo{pages}{1515--1530}.
\newblock


\bibitem[Sharma et~al\mbox{.}(2024)]%
        {socialNet2}
\bibfield{author}{\bibinfo{person}{Kartik Sharma}, \bibinfo{person}{Yeon-Chang Lee}, \bibinfo{person}{Sivagami Nambi}, \bibinfo{person}{Aditya Salian}, \bibinfo{person}{Shlok Shah}, \bibinfo{person}{Sang-Wook Kim}, {and} \bibinfo{person}{Srijan Kumar}.} \bibinfo{year}{2024}\natexlab{}.
\newblock \showarticletitle{A survey of graph neural networks for social recommender systems}.
\newblock \bibinfo{journal}{\emph{Comput. Surveys}} \bibinfo{volume}{56}, \bibinfo{number}{10} (\bibinfo{year}{2024}), \bibinfo{pages}{1--34}.
\newblock


\bibitem[Sun et~al\mbox{.}(2022)]%
        {Sun2022AdversarialAD}
\bibfield{author}{\bibinfo{person}{Lichao Sun}, \bibinfo{person}{Yingtong Dou}, \bibinfo{person}{Carl Yang}, \bibinfo{person}{Kai Zhang}, \bibinfo{person}{Ji Wang}, \bibinfo{person}{Philip~S. Yu}, \bibinfo{person}{Lifang He}, {and} \bibinfo{person}{Bo Li}.} \bibinfo{year}{2022}\natexlab{}.
\newblock \showarticletitle{Adversarial Attack and Defense on Graph Data: A Survey}.
\newblock \bibinfo{journal}{\emph{IEEE Transactions on Knowledge and Data Engineering}} (\bibinfo{year}{2022}).
\newblock


\bibitem[Sutton et~al\mbox{.}(1999)]%
        {sutton1999policy}
\bibfield{author}{\bibinfo{person}{Richard~S Sutton}, \bibinfo{person}{David McAllester}, \bibinfo{person}{Satinder Singh}, {and} \bibinfo{person}{Yishay Mansour}.} \bibinfo{year}{1999}\natexlab{}.
\newblock \showarticletitle{Policy gradient methods for reinforcement learning with function approximation}.
\newblock \bibinfo{journal}{\emph{Advances in neural information processing systems}}  \bibinfo{volume}{12} (\bibinfo{year}{1999}).
\newblock


\bibitem[Szegedy et~al\mbox{.}(2013)]%
        {Szegedy2013IntriguingPO}
\bibfield{author}{\bibinfo{person}{Christian Szegedy}, \bibinfo{person}{Wojciech Zaremba}, \bibinfo{person}{Ilya Sutskever}, \bibinfo{person}{Joan Bruna}, \bibinfo{person}{Dumitru Erhan}, \bibinfo{person}{Ian Goodfellow}, {and} \bibinfo{person}{Rob Fergus}.} \bibinfo{year}{2013}\natexlab{}.
\newblock \showarticletitle{Intriguing Properties of Neural Networks}.
\newblock \bibinfo{journal}{\emph{arXiv preprint arXiv:1312.6199}} (\bibinfo{year}{2013}).
\newblock


\bibitem[Tan et~al\mbox{.}(2022)]%
        {hgnnFinance3}
\bibfield{author}{\bibinfo{person}{Jinghua Tan}, \bibinfo{person}{Qing Li}, \bibinfo{person}{Jun Wang}, {and} \bibinfo{person}{Junxiao Chen}.} \bibinfo{year}{2022}\natexlab{}.
\newblock \showarticletitle{FinHGNN: A conditional heterogeneous graph learning to address relational attributes for stock predictions}.
\newblock \bibinfo{journal}{\emph{Information Sciences}}  \bibinfo{volume}{618} (\bibinfo{year}{2022}), \bibinfo{pages}{317--335}.
\newblock


\bibitem[Wang et~al\mbox{.}(2024)]%
        {adversarialattackHetero2}
\bibfield{author}{\bibinfo{person}{Haosen Wang}, \bibinfo{person}{Can Xu}, \bibinfo{person}{Chenglong Shi}, \bibinfo{person}{Pengfei Zheng}, \bibinfo{person}{Shiming Zhang}, \bibinfo{person}{Minhao Cheng}, {and} \bibinfo{person}{Hongyang Chen}.} \bibinfo{year}{2024}\natexlab{}.
\newblock \showarticletitle{Unsupervised Heterogeneous Graph Rewriting Attack via Node Clustering}. In \bibinfo{booktitle}{\emph{Proceedings of the 30th ACM SIGKDD Conference on Knowledge Discovery and Data Mining}}. \bibinfo{pages}{3057--3068}.
\newblock


\bibitem[Wang and Zhong(2022)]%
        {predictionGene}
\bibfield{author}{\bibinfo{person}{Li Wang} {and} \bibinfo{person}{Cheng Zhong}.} \bibinfo{year}{2022}\natexlab{}.
\newblock \showarticletitle{gGATLDA: lncRNA-disease association prediction based on graph-level graph attention network}.
\newblock \bibinfo{journal}{\emph{BMC bioinformatics}}  \bibinfo{volume}{23} (\bibinfo{year}{2022}), \bibinfo{pages}{1--24}.
\newblock


\bibitem[Wang et~al\mbox{.}(2019)]%
        {wang2019heterogeneous}
\bibfield{author}{\bibinfo{person}{Xiao Wang}, \bibinfo{person}{Houye Ji}, \bibinfo{person}{Chuan Shi}, \bibinfo{person}{Bai Wang}, \bibinfo{person}{Yanfang Ye}, \bibinfo{person}{Peng Cui}, {and} \bibinfo{person}{Philip~S Yu}.} \bibinfo{year}{2019}\natexlab{}.
\newblock \showarticletitle{Heterogeneous graph attention network}. In \bibinfo{booktitle}{\emph{The world wide web conference}}. \bibinfo{pages}{2022--2032}.
\newblock


\bibitem[Wang et~al\mbox{.}(2021)]%
        {knowledgegraphgnn3}
\bibfield{author}{\bibinfo{person}{Yu Wang}, \bibinfo{person}{Zhiwei Liu}, \bibinfo{person}{Ziwei Fan}, \bibinfo{person}{Lichao Sun}, {and} \bibinfo{person}{Philip~S Yu}.} \bibinfo{year}{2021}\natexlab{}.
\newblock \showarticletitle{Dskreg: Differentiable sampling on knowledge graph for recommendation with relational gnn}. In \bibinfo{booktitle}{\emph{Proceedings of the 30th ACM International Conference on Information \& Knowledge Management}}. \bibinfo{pages}{3513--3517}.
\newblock


\bibitem[Xiang et~al\mbox{.}(2022)]%
        {hgnnFinance2}
\bibfield{author}{\bibinfo{person}{Sheng Xiang}, \bibinfo{person}{Dawei Cheng}, \bibinfo{person}{Chencheng Shang}, \bibinfo{person}{Ying Zhang}, {and} \bibinfo{person}{Yuqi Liang}.} \bibinfo{year}{2022}\natexlab{}.
\newblock \showarticletitle{Temporal and heterogeneous graph neural network for financial time series prediction}. In \bibinfo{booktitle}{\emph{Proceedings of the 31st ACM international conference on information \& knowledge management}}. \bibinfo{pages}{3584--3593}.
\newblock


\bibitem[Xu et~al\mbox{.}(2022)]%
        {HGNNandFinance}
\bibfield{author}{\bibinfo{person}{Cong Xu}, \bibinfo{person}{Huiling Huang}, \bibinfo{person}{Xiaoting Ying}, \bibinfo{person}{Jianliang Gao}, \bibinfo{person}{Zhao Li}, \bibinfo{person}{Peng Zhang}, \bibinfo{person}{Jie Xiao}, \bibinfo{person}{Jiarun Zhang}, {and} \bibinfo{person}{Jiangjian Luo}.} \bibinfo{year}{2022}\natexlab{}.
\newblock \showarticletitle{HGNN: Hierarchical graph neural network for predicting the classification of price-limit-hitting stocks}.
\newblock \bibinfo{journal}{\emph{Information Sciences}}  \bibinfo{volume}{607} (\bibinfo{year}{2022}), \bibinfo{pages}{783--798}.
\newblock


\bibitem[Yang and Han(2023)]%
        {hgnnCitation3}
\bibfield{author}{\bibinfo{person}{Carl Yang} {and} \bibinfo{person}{Jiawei Han}.} \bibinfo{year}{2023}\natexlab{}.
\newblock \showarticletitle{Revisiting citation prediction with cluster-aware text-enhanced heterogeneous graph neural networks}. In \bibinfo{booktitle}{\emph{2023 IEEE 39th International Conference on Data Engineering (ICDE)}}. IEEE, \bibinfo{pages}{682--695}.
\newblock


\bibitem[Ye et~al\mbox{.}(2022)]%
        {gnnandknowgraph}
\bibfield{author}{\bibinfo{person}{Zi Ye}, \bibinfo{person}{Yogan~Jaya Kumar}, \bibinfo{person}{Goh~Ong Sing}, \bibinfo{person}{Fengyan Song}, {and} \bibinfo{person}{Junsong Wang}.} \bibinfo{year}{2022}\natexlab{}.
\newblock \showarticletitle{A comprehensive survey of graph neural networks for knowledge graphs}.
\newblock \bibinfo{journal}{\emph{IEEE Access}}  \bibinfo{volume}{10} (\bibinfo{year}{2022}), \bibinfo{pages}{75729--75741}.
\newblock


\bibitem[Zhang et~al\mbox{.}(2022)]%
        {zhang2022robust}
\bibfield{author}{\bibinfo{person}{Mengmei Zhang}, \bibinfo{person}{Xiao Wang}, \bibinfo{person}{Meiqi Zhu}, \bibinfo{person}{Chuan Shi}, \bibinfo{person}{Zhiqiang Zhang}, {and} \bibinfo{person}{Jun Zhou}.} \bibinfo{year}{2022}\natexlab{}.
\newblock \showarticletitle{Robust heterogeneous graph neural networks against adversarial attacks}. In \bibinfo{booktitle}{\emph{Proceedings of the AAAI Conference on Artificial Intelligence}}, Vol.~\bibinfo{volume}{36}. \bibinfo{pages}{4363--4370}.
\newblock


\bibitem[Zhao et~al\mbox{.}(2019)]%
        {hgnnCitation2}
\bibfield{author}{\bibinfo{person}{Fen Zhao}, \bibinfo{person}{Yi Zhang}, \bibinfo{person}{Jianguo Lu}, {and} \bibinfo{person}{Ofer Shai}.} \bibinfo{year}{2019}\natexlab{}.
\newblock \showarticletitle{Measuring academic influence using heterogeneous author-citation networks}.
\newblock \bibinfo{journal}{\emph{Scientometrics}}  \bibinfo{volume}{118} (\bibinfo{year}{2019}), \bibinfo{pages}{1119--1140}.
\newblock


\bibitem[Zhao et~al\mbox{.}(2024a)]%
        {HGAttack}
\bibfield{author}{\bibinfo{person}{He Zhao}, \bibinfo{person}{Zhiwei Zeng}, \bibinfo{person}{Yongwei Wang}, \bibinfo{person}{Deheng Ye}, {and} \bibinfo{person}{Chunyan Miao}.} \bibinfo{year}{2024}\natexlab{a}.
\newblock \showarticletitle{HGAttack: Transferable Heterogeneous Graph Adversarial Attack}.
\newblock \bibinfo{journal}{\emph{arXiv preprint arXiv:2401.09945}} (\bibinfo{year}{2024}).
\newblock


\bibitem[Zhao et~al\mbox{.}(2024b)]%
        {zhao2024hgattack}
\bibfield{author}{\bibinfo{person}{He Zhao}, \bibinfo{person}{Zhiwei Zeng}, \bibinfo{person}{Yongwei Wang}, \bibinfo{person}{Deheng Ye}, {and} \bibinfo{person}{Chunyan Miao}.} \bibinfo{year}{2024}\natexlab{b}.
\newblock \showarticletitle{HGAttack: Transferable Heterogeneous Graph Adversarial Attack}.
\newblock \bibinfo{journal}{\emph{arXiv preprint arXiv:2401.09945}} (\bibinfo{year}{2024}).
\newblock


\bibitem[Zhu et~al\mbox{.}(2024)]%
        {adversarialattackGraph}
\bibfield{author}{\bibinfo{person}{Guanghui Zhu}, \bibinfo{person}{Mengyu Chen}, \bibinfo{person}{Chunfeng Yuan}, {and} \bibinfo{person}{Yihua Huang}.} \bibinfo{year}{2024}\natexlab{}.
\newblock \showarticletitle{Simple and efficient partial graph adversarial attack: A new perspective}.
\newblock \bibinfo{journal}{\emph{IEEE Transactions on Knowledge and Data Engineering}} (\bibinfo{year}{2024}).
\newblock


\bibitem[Zhu et~al\mbox{.}(2019)]%
        {RHGNN}
\bibfield{author}{\bibinfo{person}{Shichao Zhu}, \bibinfo{person}{Chuan Zhou}, \bibinfo{person}{Shirui Pan}, \bibinfo{person}{Xingquan Zhu}, {and} \bibinfo{person}{Bin Wang}.} \bibinfo{year}{2019}\natexlab{}.
\newblock \showarticletitle{Relation structure-aware heterogeneous graph neural network}. In \bibinfo{booktitle}{\emph{2019 IEEE international conference on data mining (ICDM)}}. IEEE, \bibinfo{pages}{1534--1539}.
\newblock


\bibitem[Zhu et~al\mbox{.}(2020)]%
        {HGCN}
\bibfield{author}{\bibinfo{person}{Zhihua Zhu}, \bibinfo{person}{Xinxin Fan}, \bibinfo{person}{Xiaokai Chu}, {and} \bibinfo{person}{Jingping Bi}.} \bibinfo{year}{2020}\natexlab{}.
\newblock \showarticletitle{HGCN: A heterogeneous graph convolutional network-based deep learning model toward collective classification}. In \bibinfo{booktitle}{\emph{Proceedings of the 26th ACM SIGKDD international conference on knowledge discovery \& data mining}}. \bibinfo{pages}{1161--1171}.
\newblock


\bibitem[Z{\"u}gner et~al\mbox{.}(2020)]%
        {AdversarialAttacksonGraphNeuralNetworks}
\bibfield{author}{\bibinfo{person}{Daniel Z{\"u}gner}, \bibinfo{person}{Oliver Borchert}, \bibinfo{person}{Amir Akbarnejad}, {and} \bibinfo{person}{Stephan G{\"u}nnemann}.} \bibinfo{year}{2020}\natexlab{}.
\newblock \showarticletitle{Adversarial attacks on graph neural networks: Perturbations and their patterns}.
\newblock \bibinfo{journal}{\emph{ACM Transactions on Knowledge Discovery from Data (TKDD)}} \bibinfo{volume}{14}, \bibinfo{number}{5} (\bibinfo{year}{2020}), \bibinfo{pages}{1--31}.
\newblock


\end{thebibliography}

\newpage

\appendix
\section{Workflow of HeteroKRLAttack}
\label{appendix:algorithm_workflow}
\begin{algorithm}[h]
\caption{HeteroKRLAttack: Training and Inference}
\label{alg:heterokrlattack}
\begin{algorithmic}[1]

\State \textcolor{gray}{// \textbf{Training Phase}}
\State \textbf{Input:} Black-box model $f$, graph $G$, target node $v$, budget $c$, hyperparameters $(\gamma, \alpha)$
\State Initialize policy $\pi_\theta$ (TypeNet + ActionNet), trajectory buffer $D$, set $G_{\text{cur}} \gets G$
\For{$t = 1$ to $c$}
    \State $s \gets (v, V_{\text{obs}})$
    \If{$t$ is even} 
        \State Select auxiliary node type $t_a \sim \text{TypeNet}(s)$
    \Else 
        \State Select node $v_a \sim \text{ActionNet}(s, t_a)$
        \State $G_{\text{cur}} \gets G_{\text{cur}} \oplus (v, v_a)$
        \State Update reward $r$ according to Equation~(3)
        \State Append $(s, (v, v_a), r)$ to $D$
        \If{$r$ is a success signal} \textbf{break} \EndIf
    \EndIf
\EndFor

\State \textcolor{gray}{// Update policy via REINFORCE}
\ForAll{$(s, a, r) \in D$}
    \State Compute return $R_t = \sum_{k \ge t} \gamma^{k-t} r_k$
    \State $\theta \gets \theta + \alpha \nabla_\theta \log \pi_\theta(a|s) \cdot R_t$
\EndFor

\Statex \textcolor{gray}{// \textbf{Inference Phase}}
\State \textbf{Input:} Trained policy $\pi_\theta$, graph $G$, target node $v$, budget $c$, Top-$K$ parameter $K$
\State $G_{\text{cur}}\gets G$
\For{$t=1$ to $c$}
  \State $s\gets(v,\,V_{\mathrm{obs}})$
  \State $(t_a,\,v_0)\sim\pi_\theta(s)$  \Comment{sample type then node}
  \State $S_K\gets\mathrm{KDTreeQuery}(\mathrm{feat}(v_0),\,K)$  
  \State $v^*\gets\arg\max_{u\in S_K}L\bigl(f(v\mid G_{\text{cur}}\oplus(v,u)),y_{\text{true}}\bigr)$ 
  \State $G_{\text{cur}}\gets G_{\text{cur}}\oplus(v,\,v^*)$
\EndFor
\State \Return $G_{\text{cur}}$
\end{algorithmic}
\end{algorithm}

\section{Time Complexity}
\label{appendix:time_complexity}
In this section, we provide a detailed analysis of the time complexity for our proposed HeteroKRLAttack framework, covering both the training and inference phases. For this analysis, we define several key variables: $c$ represents the perturbation budget; $|\mathcal{T}_{\text{aux}}|$ is the number of auxiliary node types; $N_{\text{max\_type}}$ denotes the maximum number of nodes within any single auxiliary type; $d$ is the feature dimension of the nodes; $K$ is a hyperparameter representing the proportion of candidate nodes selected for refinement; $T_f$ is the time for a single query to the black-box victim model $f(\cdot)$; and $T_{\text{ResMLP}}$ is the complexity of a single forward pass through the residual MLP in ActionNet.

The training phase complexity is determined by two main stages: trajectory generation and policy updates. During trajectory generation, the agent interacts with the environment for up to $c$ steps. Each step involves selecting an auxiliary type via TypeNet, which is computationally negligible, followed by selecting a node via ActionNet, which has a worst-case complexity of $O(N_{\text{max\_type}} \cdot T_{\text{ResMLP}})$. The reward calculation is dominated by a black-box query costing $T_f$. Thus, generating a full trajectory costs $O(c \cdot (N_{\text{max\_type}} \cdot T_{\text{ResMLP}} + T_f))$. Subsequently, the policy update using the REINFORCE algorithm requires iterating through the trajectory, resulting in a complexity of $O(c \cdot (c + N_{\text{max\_type}} \cdot T_{\text{ResMLP}}))$. Combining these stages, the overall time complexity for one training episode is:
\[
O(c \cdot (N_{\text{max\_type}} \cdot T_{\text{ResMLP}} + T_f) + c^2)
\]
The dominant computational costs arise from the evaluation of all candidate nodes in ActionNet and the repeated queries to the black-box model.

During the inference phase, the trained agent performs an attack over $c$ steps, enhanced by the Top-K refinement mechanism. In each step, the agent first selects an initial candidate node with a complexity of $O(N_{\text{max\_type}} \cdot T_{\text{ResMLP}})$. The core of the process is the Top-K refinement, which begins with constructing a KD-Tree on the node features of the selected auxiliary type, costing $O(N_{\text{max\_type}} \cdot d \cdot \log N_{\text{max\_type}})$ in the worst case. The most significant bottleneck follows: the framework evaluates a refined set of candidates to find the most disruptive node. The number of candidates is $K \cdot N_{\text{max\_type}}$, and evaluating each requires a black-box query. This step's complexity is $O((K \cdot N_{\text{max\_type}}) \cdot T_f)$. Summing the costs for a single modification and multiplying by the budget $c$, the total time complexity for the inference phase is:
\[
O(c \cdot (N_{\text{max\_type}} \cdot d \cdot \log N_{\text{max\_type}} + (K \cdot N_{\text{max\_type}}) \cdot T_f))
\]
This highlights that the inference cost is heavily dependent on the size of the node sets, the feature dimensionality, and, most critically, the number of black-box queries dictated by the hyperparameter $K$.

\section{Experiment Settings}
\label{appendix: experiment_victim_settings}
\begin{table}[h]
\centering
\caption{Victim Model Training Hyperparameters}
\label{tab:victim_hyperparameters}
\small
\begin{tabular}{lccc}
\hline
\textbf{Hyperparameter} & \textbf{HAN} & \textbf{HGT} & \textbf{SimpleHGN} \\
\hline
Number of Layers        & 2            & 8            & 5 \\
Hidden Dimension        & 256          & 64           & 512 \\
Dropout Rate            & 0.2          & 0.2          & 0.1 \\
Learning Rate           & 0.005        & 0.001        & 0.001 \\
Training Epochs         & 400          & 400          & 400 \\
Optimizer               & \multicolumn{3}{c}{Adam} \\
LR Scheduler            & \multicolumn{3}{c}{OneCycleLR} \\
Loss Function           & \multicolumn{3}{c}{Cross-Entropy} \\
Early Stopping          & \multicolumn{3}{c}{Patience = 50} \\
\hline
\end{tabular}
\end{table}

\end{document}